\documentclass[11pt,a4paper]{article}
\usepackage[hyperref]{emnlp2020}
\usepackage{times}
\usepackage{comment}
\usepackage{latexsym}
\usepackage{multirow,bigstrut}
\usepackage{graphicx}
\usepackage{subcaption}
\usepackage{array}
\usepackage{float}
\graphicspath{ {./images/} }

\usepackage{microtype}

\aclfinalcopy 
\def\aclpaperid{1280} 

\title{{P}retrained Language Model Embryology: {T}he Birth of {ALBERT}}

\author{Cheng-Han Chiang \\
  National Taiwan University,\\ Taiwan\\
  \texttt{dcml0714@gmail.com} \\\And
   Sung-Feng Huang \\
   National Taiwan University,\\ Taiwan\\
   \texttt{f06942045@ntu.edu.tw} \\\And
   Hung-yi Lee \\
  National Taiwan University,\\ Taiwan \\
  \texttt{hungyilee@ntu.edu.tw} \\}

\date{}

\begin{document}
\maketitle
\begin{abstract}
While behaviors of pretrained language models (LMs) have been thoroughly examined, what happened during pretraining is rarely studied.
We thus investigate the developmental process from a set of randomly initialized parameters to a totipotent\footnote{According to Wikipedia, totipotency is the ability of a single cell to divide and produce all of the differentiated cells in an organism. We use its adjective form here to refer to the ability of a pretrained model which can be finetuned for a variety of downstream tasks.} language model, which we refer to as the \textit{embryology} of a pretrained language model.
Our results show that ALBERT learns to reconstruct and predict tokens of different parts of speech (POS) in different learning speeds during pretraining.
We also find that linguistic knowledge and world knowledge do not generally improve as pretraining proceeds, nor do downstream tasks' performance.
These findings suggest that knowledge of a pretrained model varies during pretraining, and having more pretrain steps does not necessarily provide a model with more comprehensive knowledge. 
We provide source codes and pretrained models to reproduce our results at \url{https://github.com/d223302/albert-embryology}.

\end{abstract}


\section{Introduction}

The world of NLP has gone through some tremendous revolution since the proposal of contextualized word embeddings. 
Some big names are ELMo \citep{Peters:2018}, GPT \citep{radford2019language}, and BERT ~\citep{devlin2018bert}, along with its variants ~\citep{sanh2019distilbert, liu2019roberta, lan2019albert}. 
Performance boosts on miscellaneous downstream tasks have been reported by finetuning these totipotent pretrained language models.
With a view to better grasping what has been learned by these contextualized word embedding models, probing is generally applied to the pretrained models and the models finetuned from them.
Probing targets can range from linguistic knowledge, including semantic roles and syntactic structures ~\citep{liu2019linguistic, tenney2019bert, tenney2019you, hewitt2019structural}, to world knowledge ~\citep{petroni2019language}.

While the previous work focuses on what knowledge has been learned after pretraining of transformer-based language models, few delve into their dynamics during pretraining. 
What happened during the training process of a deep neural network model has been widely studied, including ~\citet{gur2018gradient}, ~\citet{frankle2019early}, ~\citet{raghu2017svcca}, ~\citet{morcos2018insights}. 
Some previous works also study the dynamics of the training process of an LSTM language model~\citep{saphra2018language, saphra2019understanding}, but the training dynamics of a large scale pretrained language models are not well-studied.
In this work, we probe ALBERT \citep{lan2019albert} during its pretraining phase every \(N\) parameter update steps and study what it has learned and what it can achieve so far.
We perform a series of experiments, detailed in the following sections, to investigate the development of predicting and reconstructing tokens (Section~\ref{section: mask}), how linguistic and world knowledge evolve through time (Section~\ref{sec:probe}, Section~\ref{section: knowledge}), and whether amassing those information serves as an assurance of good downstream task performances (Section~\ref{section: downstream}). 

We have the following findings based on ALBERT:
\begin{itemize}
\item The prediction and reconstruction of tokens with different POS tags have different learning speeds. (Section~\ref{section: mask})
\item Semantic and syntactic knowledge is developed simultaneously in ALBERT. (Section~\ref{sec:probe})
\item Finetuning from model pretrained for 250k steps gives a decent GLUE score (80.23), and further pretrain steps only make the GLUE score rise as high as 81.50. 
\item While ALBERT does generally gain more world knowledge as pretraining goes on, the model seems to be dynamically renewing its knowledge about the world. (Section~\ref{section: knowledge})
\end{itemize}

While we only include the detailed results of ALBERT in the main text, we find that the results also generalize to the other two transformer-based language models, ELECTRA ~\citep{clark2019electra} and BERT, which are quite different from ALBERT in the sense of pretext task and model architecture.
We put the detailed results of ELECTRA and BERT in the appendix.

\section{Pretraining ALBERT}
\label{sec:pretraining}
ALBERT is a variant of BERT with cross-layer parameters sharing and factorized embedding parameterization. 
The reason why we initially chose ALBERT as our subject lies in its parameter efficiency, which becomes a significant issue when we need to store 1000 checkpoints during the pretraining process.

To investigate what happened during the pretraining process of ALBERT, we pretrained an ALBERT-base model ourselves.
To maximally reproduce the results in \citet{lan2019albert}, we follow most of the training hyperparameters in the original work, only modifying some hyperparameters to fit in our limited computation resources\footnote{We use the official implementation of ALBERT at \url{https://github.com/google-research/albert}.}.
We also follow \citet{lan2019albert}, using English Wikipedia as our pretraining data, and we use the Project Guttenberg Dataset \citep{lahiri:2014:SRW} instead of \textsc{B}ook\textsc{C}orpus.
The total size of the corpus used in pretraining is 16GB.
The pretraining was done on a single Cloud TPU V3 and took eight days to finish 1M pretrain steps, costing around 700 USD.
More details on pretraining are specified in appendix~\ref{app: pretrain hpp}.

\begin{figure}[t]
\centering
\begin{subfigure}[b]{0.45\textwidth}
\centering
\includegraphics[clip, trim = 15px 2px 39px
40px,width=\linewidth]{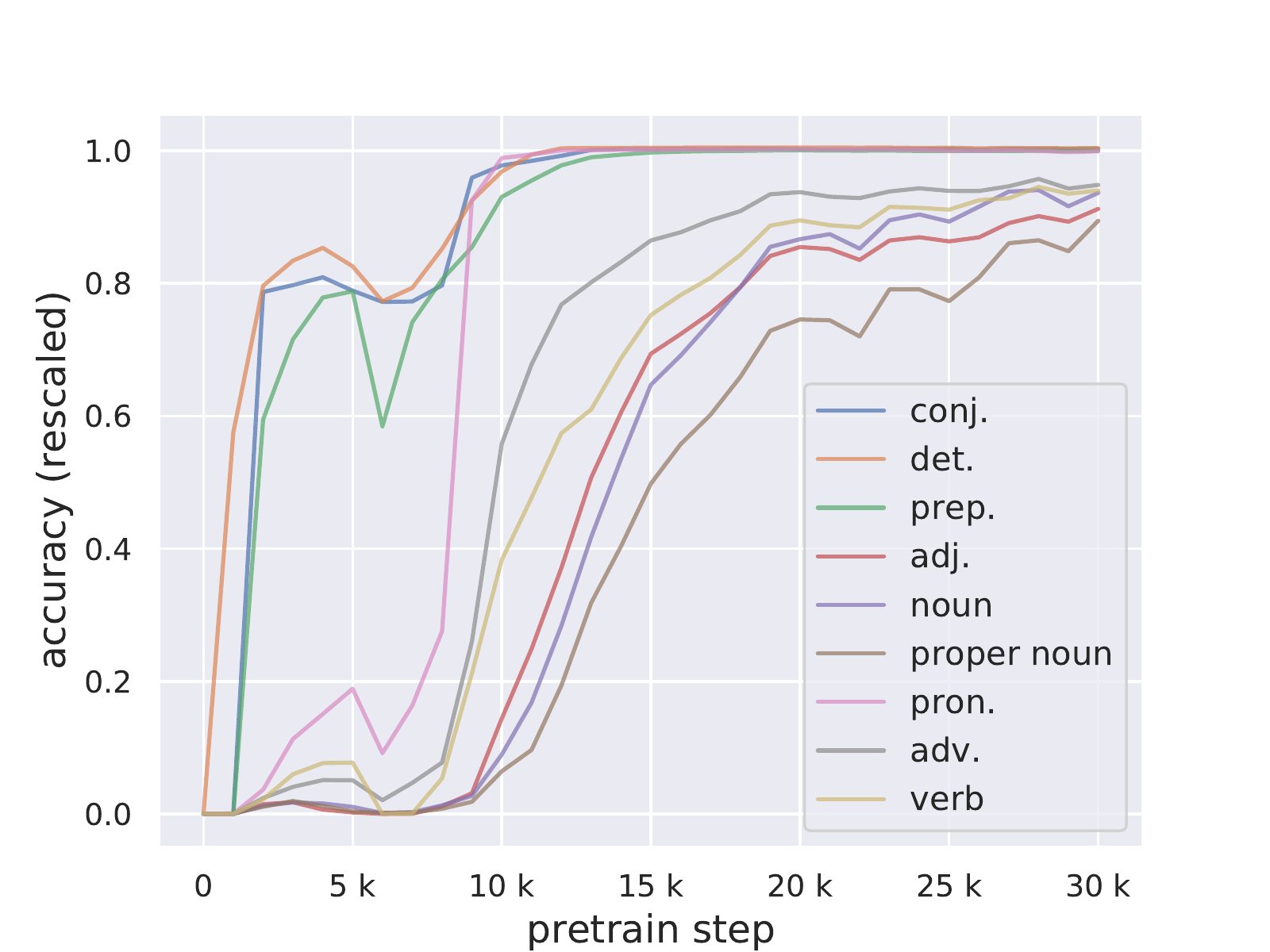}
\caption{Token reconstruction}
\label{fig:recon}
\end{subfigure}
\begin{subfigure}[b]{0.45\textwidth}
\centering
\includegraphics[clip, trim = 15px 2px 39px 40px,width=\linewidth]{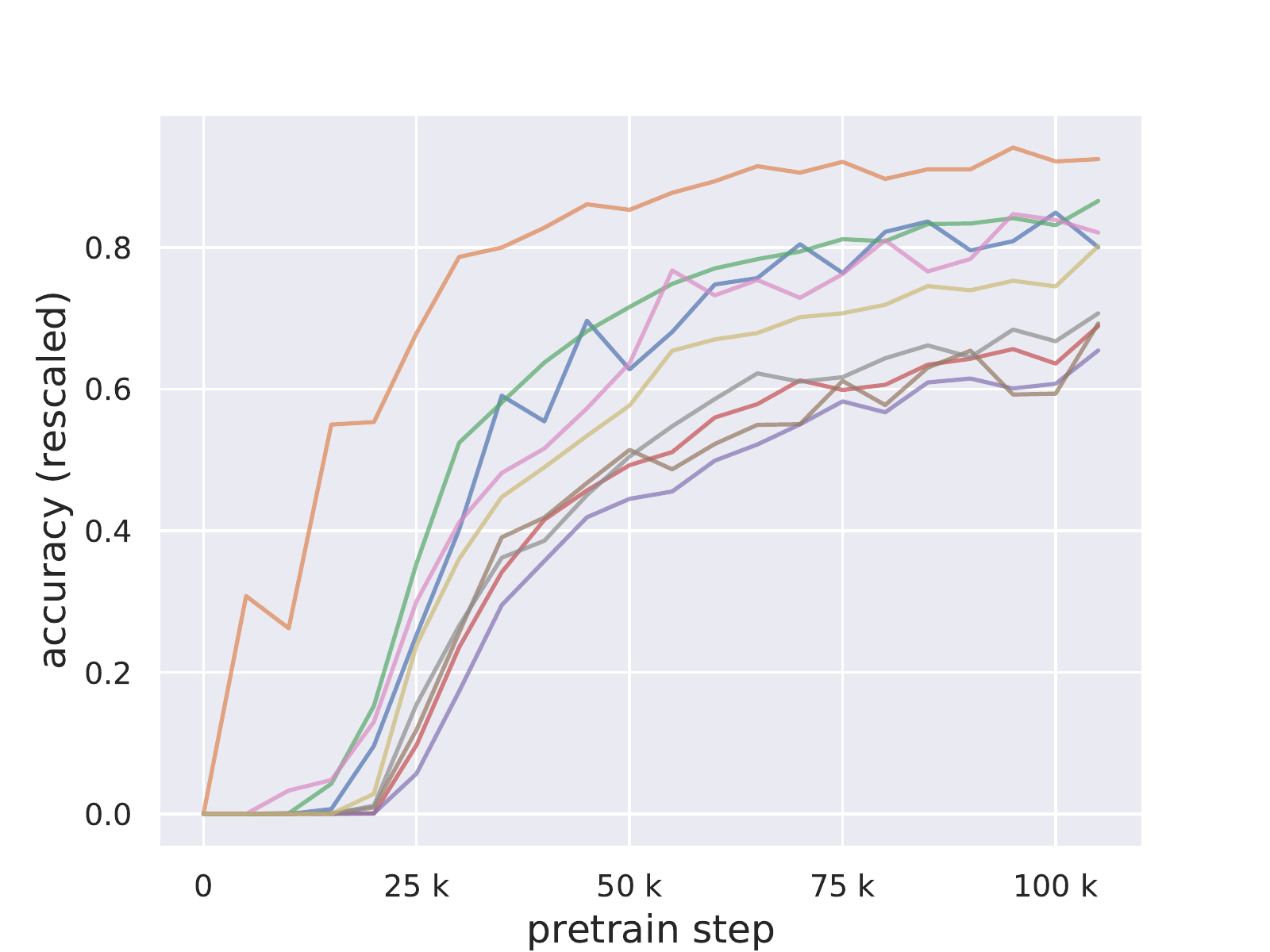}
\caption{Mask prediction}
\label{fig:mask_pred}
\end{subfigure}
\caption{Rescaled accuracy of token reconstruction and mask prediction during pretraining. We rescale the accuracy of each line by the accuracy when the model is fully pretrained, i.e., the accuracy after pretraining 1M steps. Token reconstruction are evaluated every 1K pretrain steps, and mask prediction evaluated every 5K steps.}
\label{fig:reconstruction_and_mask_predict}
\end{figure}
\section{Learning to Predict the Masked Tokens and Reconstruct the Input Tokens}
\label{section: mask}

During the pretraining stage of a masked LM (MLM), it learns to predict masked tokens based on the remaining unmasked part of the sentence, and it also learns to reconstruct token identities of unmasked tokens from their output representations of the model.
Better prediction and reconstruction results indicate the model being able to utilize contextual information.
To maximally reconstruct the input tokens, the output representations must keep sufficient information regarding token identities.

We investigate the behavior of mask prediction and token reconstruction for tokens of different POS during the early stage of pretraining. 
We use the POS tagging in OntoNotes 5.0 \citep{2013ontonotes} in this experiment.
For the mask prediction part, we mask a whole word (which may contain multiple tokens) of an input sentence, feed the masked sentence into ALBERT, and predict the masked token(s).
We evaluate the prediction performance by calculating the prediction's accuracy based on POS of the word; the predicted token(s) should exactly match the original token(s) to be deemed an accurate prediction.
As for the token reconstruction part, the input to the model is simply the original sentence.

The results of reconstruction are shown in Figure~\ref{fig:reconstruction_and_mask_predict}(a).
ALBERT first learns to reconstruct function words, e.g., determiners, prepositions, and then gradually learns to reconstruct content words in the order of verb, adverb, adjective, noun, and proper noun. 
We also found that different forms and tenses of a verb do not share the same learning schedule, with third-person singular present being the easiest to reconstruct and present participle being the hardest (shown in Appendix~\ref{app: verb}). 
The prediction results in Figure~\ref{fig:reconstruction_and_mask_predict}(b) reveal that learning mask prediction is generally more challenging than token reconstruction.
ALBERT learns to predict masked tokens with an order similar to token reconstruction, though much slower and less accurate.
We find that BERT also learns to perform mask prediction and token reconstruction in a similar fashion, with the results provided in Appendix~\ref{app:mask_predict_bert}.

\begin{figure}[t]
\centering
\begin{subfigure}[b]{0.4\textwidth}
\centering
\includegraphics[clip, trim = 15px 2px 39px
30px,width=\linewidth]{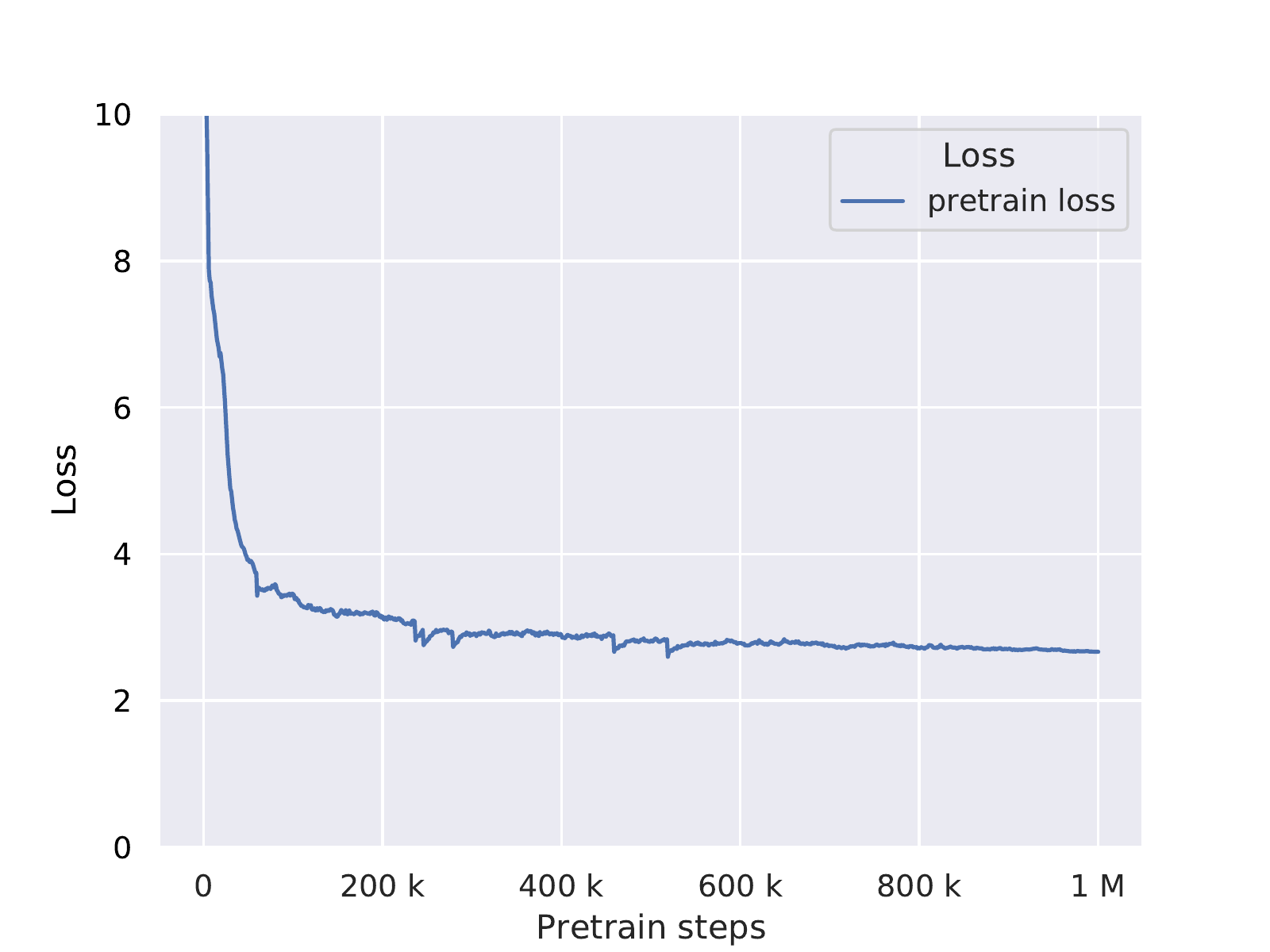}
\caption{Total loss during pretraining}
\label{fig:pretrain_loss}
\end{subfigure}
\begin{subfigure}[b]{0.4\textwidth}
\centering
\includegraphics[clip, trim = 15px 2px 39px 30px,width=\linewidth]{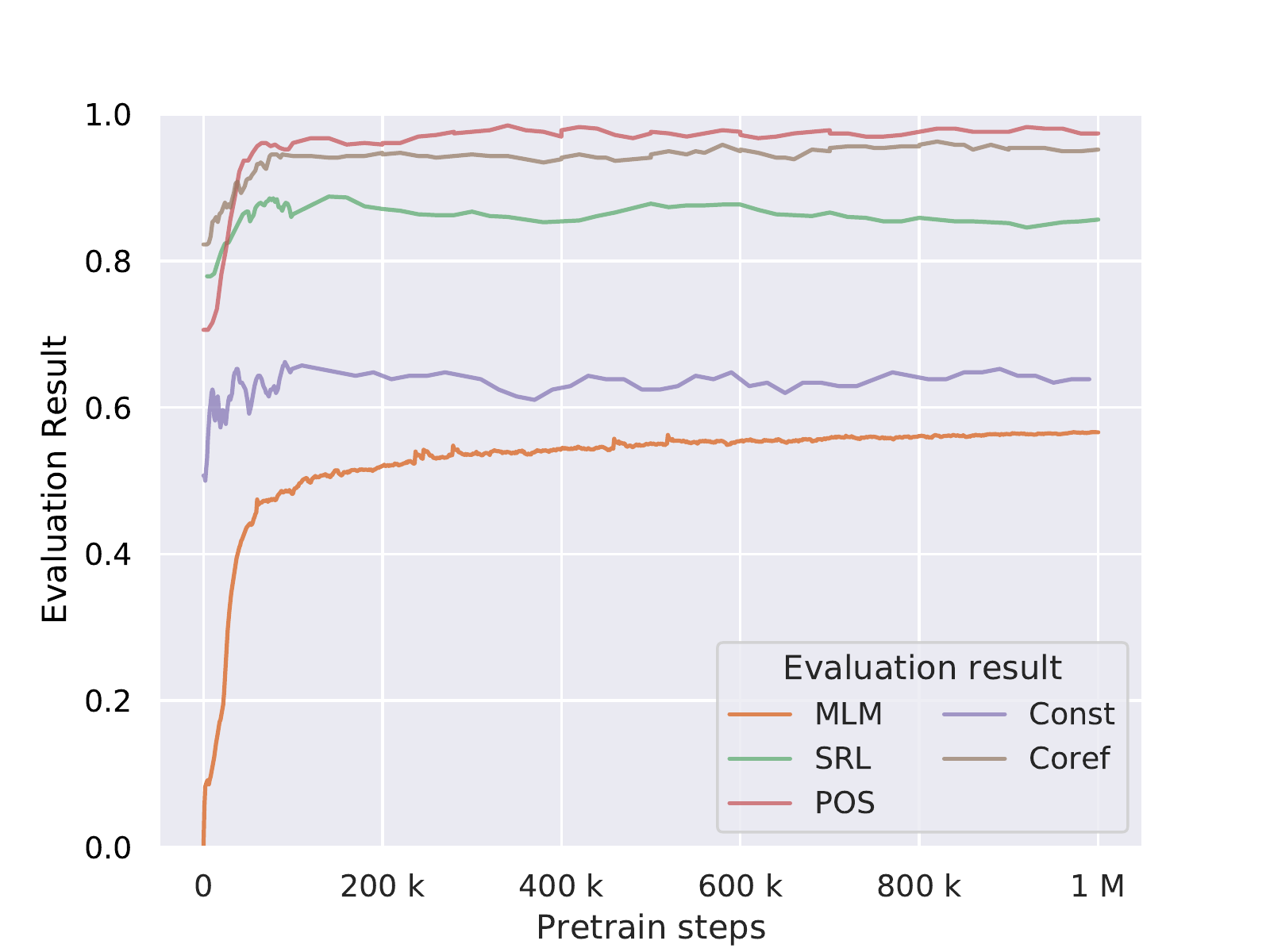}
\caption{Masked LM accuracy and F1 scores of different probing tasks over the course of pretraining}
\label{fig:pretrain_probe}
\end{subfigure}
\caption{The probing results of hidden representation from layer 8; all four tasks are evaluated with test set of OntoNotes 5.0 and F1 scores are reported. MLM accuracy is also shown. We smoothed the lines by averaging 3 consecutive data points for better illustration. The unsmoothed result is in Appendix~\ref{app: probe}.}
\label{fig:pretrain}
\end{figure}

\section{Probing Linguistic Knowledge Development During Pretraining}
\label{sec:probe}
Probing is widely used to understand what kind of information is encoded in embeddings of a language model.
In short, probing experiments train a task-specific classifier to examine if token embeddings contain the knowledge required for the probing task.
Different language models may give different results on different probing tasks, and representations from different layers of a language model may also contain different linguistic information \citep{liu2019linguistic, tenney2019you}.

Our probing experiments are modified from the ``edge probing'' framework in \citet{tenney2019you}.
\citet{hewitt2019designing} previously showed that probing models should be \textit{selective}, so we use linear classifiers for probing.
We select four probing tasks for our experiments: part of speech (POS) tagging, constituent (const) tagging, coreference (coref) resolution, and semantic role labeling (SRL).
The former two tasks probe syntactic knowledge hidden in token embeddings, and the last two tasks are designed to inspect the semantic knowledge provided by token embeddings.
We use annotations provided in OntoNotes 5.0 \citep{2013ontonotes} in our experiments.

The probing results are shown in Figure~\ref{fig:pretrain_probe}.
We observe that all four tasks show similar trends during pretraining, indicating that semantic knowledge and syntactic knowledge are developed simultaneously during pretraining.
For syntactically related tasks, the performance of both POS tagging and constituent tagging boost very fast in the first 100k pretrain steps, and no further improvement can be seen throughout the remaining pretraining process, while performance fluctuates from time to time.
We also observe an interesting phenomenon: the probed performances of SRL peak at around 150k steps and slightly decay over the remaining pretraining process, suggesting that some information in particular layers related to probing has been dwindling while the ALBERT model strives to advance its performance on the pretraining objective.
The loss of the pretraining objective is also shown in Figure~\ref{fig:pretrain_loss}.

\begin{figure}[t!]
\centering
\includegraphics[clip, trim = 15px 2px 10px
10px,width=\linewidth]{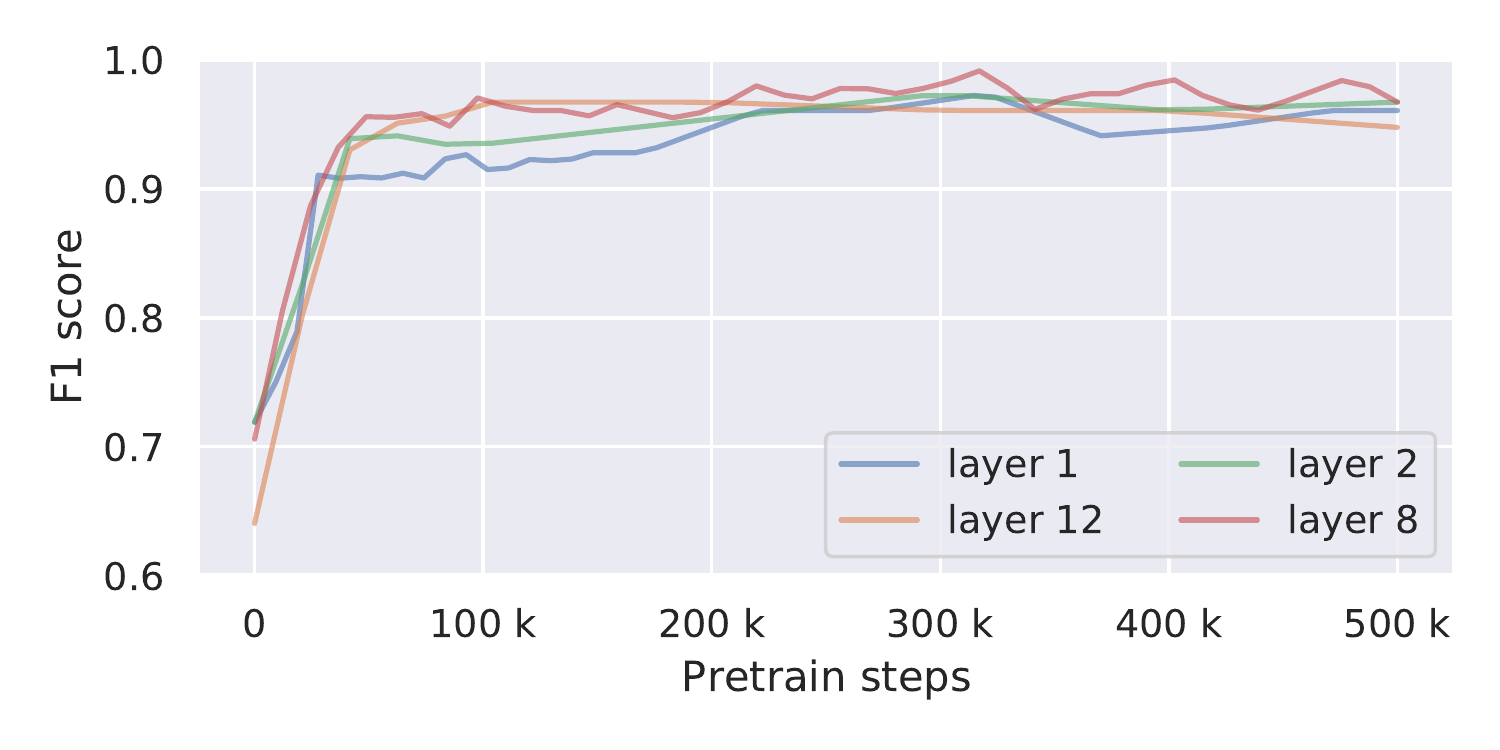}
\caption{The probing results of POS during pretraining.
Layers are indexed from the input layer to the output layer.}
\label{fig:pos}
\end{figure}

Scrutinizing the probing results of different layers (Figure~\ref{fig:pos} and Appendix~\ref{app: probe}), we find that the behaviors among different layers are slightly different. 
While the layers closer to output layer perform worse than layers closer to input layer at the beginning of pretraining, their performances rise drastically and eventually surpass the top few layers; however, they start to decay after they reach best performances.  
\begin{figure}[t]
\centering
\small
\setlength{\tabcolsep}{1pt} 
\renewcommand{\arraystretch}{0.1}
\begin{tabular}{ccccc}
    0 & 30k & 60k & 210k & 500k \\
    \includegraphics[width=0.185\linewidth, trim =30px 20px 24px 8px, clip]{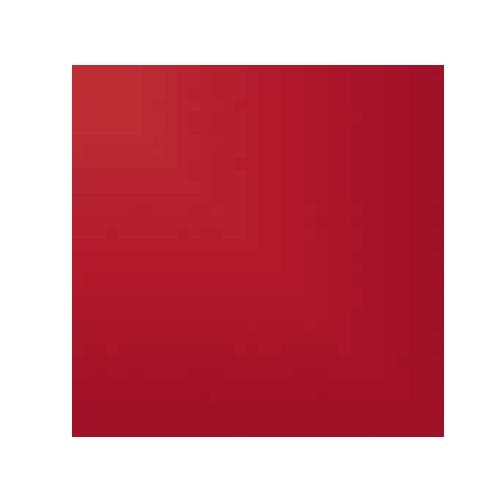} &
    \includegraphics[width=0.185\linewidth, trim =30px 20px 24px 8px, clip]{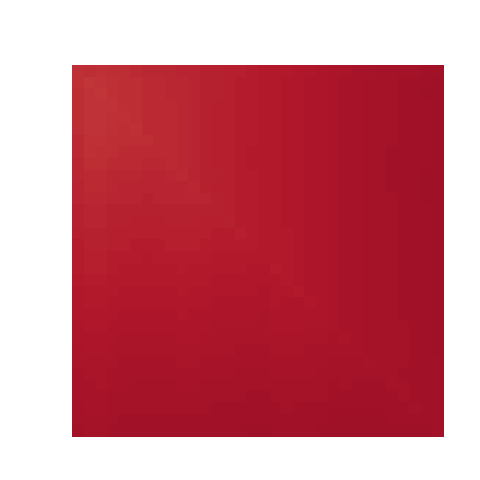} &
    \includegraphics[width=0.185\linewidth, trim =30px 20px 24px 8px, clip]{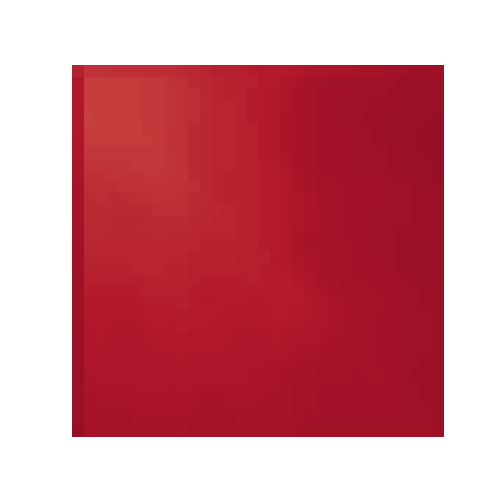} &
    \includegraphics[width=0.185\linewidth, trim =30px 20px 24px 8px, clip]{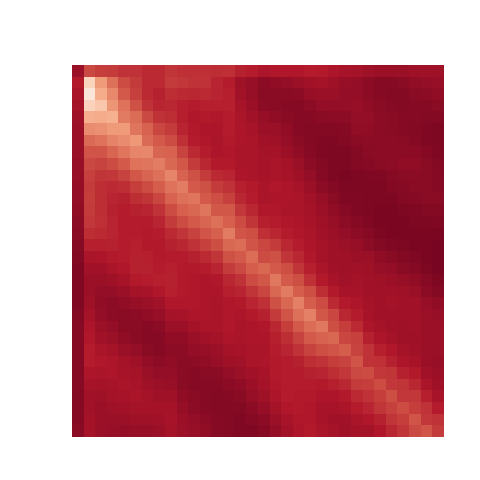} &
    \includegraphics[width=0.185\linewidth, trim =30px 20px 24px 8px, clip]{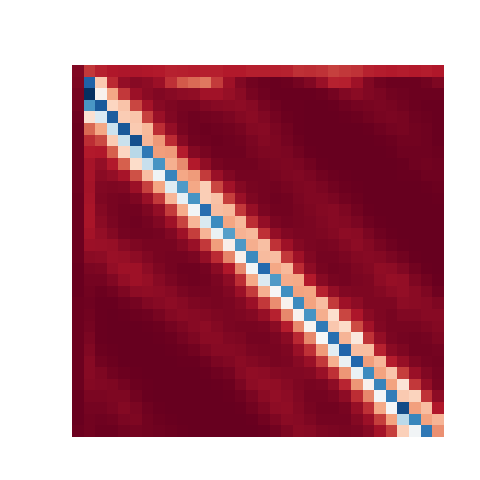} \\
    \includegraphics[width=0.185\linewidth, trim =30px 20px 24px 8px, clip]{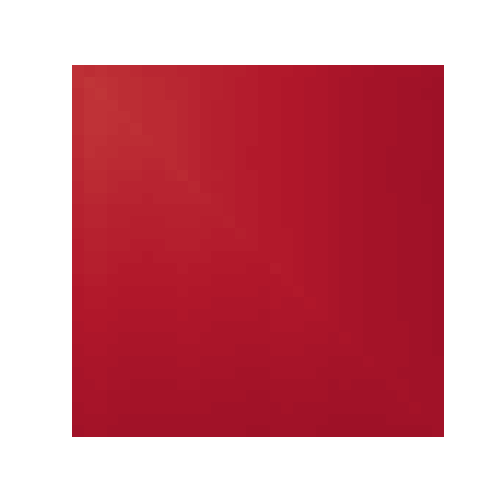} &
    \includegraphics[width=0.185\linewidth, trim =30px 20px 24px 8px, clip]{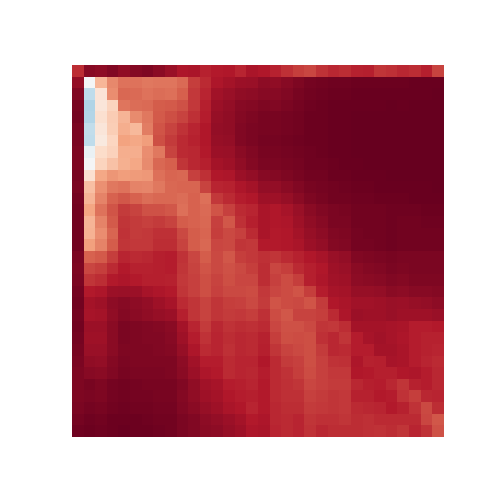} &
    \includegraphics[width=0.185\linewidth, trim =30px 20px 24px 8px, clip]{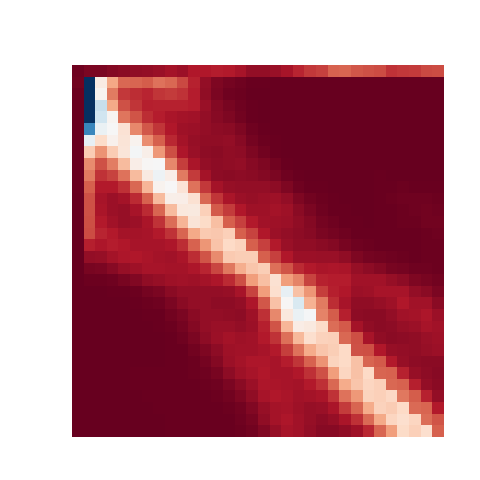} &
    \includegraphics[width=0.185\linewidth, trim =30px 20px 24px 8px, clip]{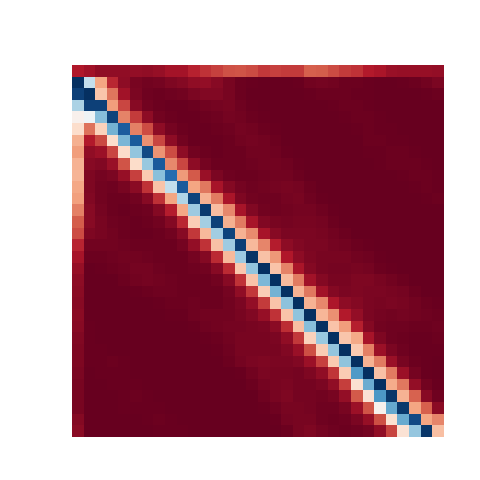} &
    \includegraphics[width=0.185\linewidth, trim =30px 20px 24px 8px, clip]{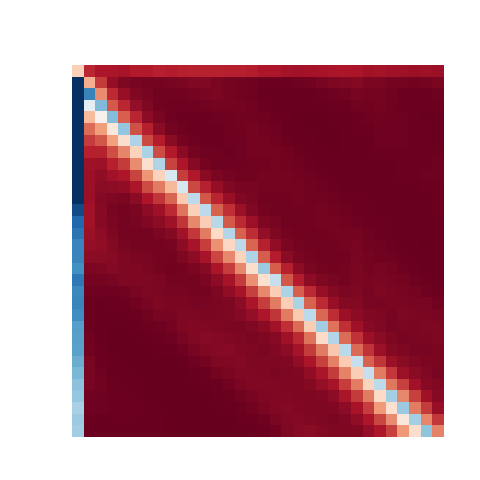} \\
    \includegraphics[width=0.185\linewidth, trim =30px 20px 24px 8px, clip]{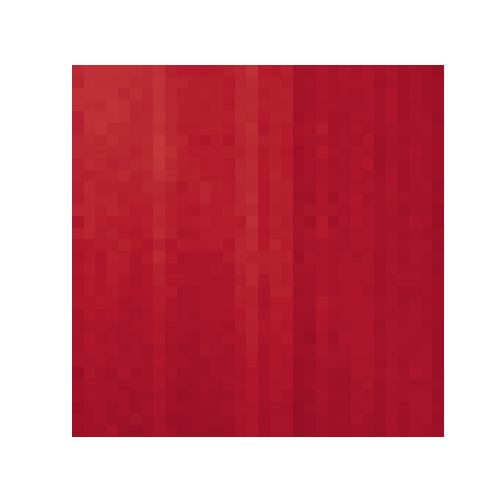} &
    \includegraphics[width=0.185\linewidth, trim =30px 20px 24px 8px, clip]{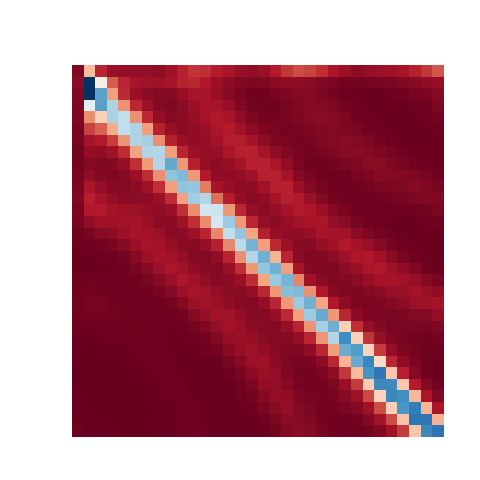} &
    \includegraphics[width=0.185\linewidth, trim =30px 20px 24px 8px, clip]{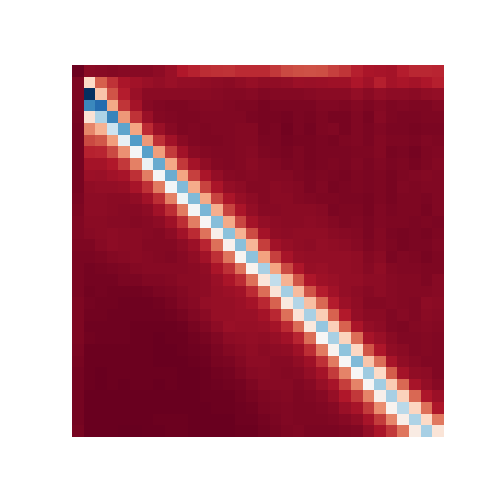} &
    \includegraphics[width=0.185\linewidth, trim =30px 20px 24px 8px, clip]{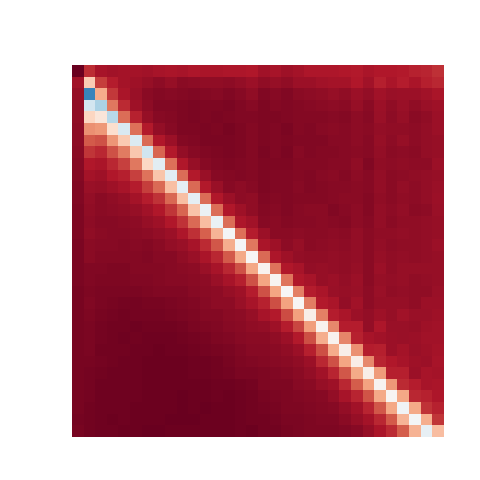} &
    \includegraphics[width=0.185\linewidth, trim =30px 20px 24px 8px, clip]{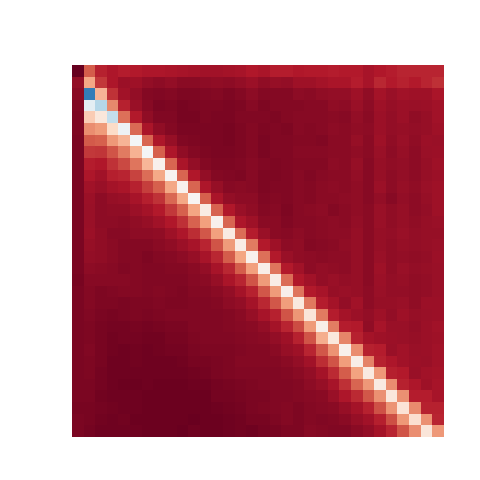} \\
    \multicolumn{5}{c}{\centering\includegraphics[scale=0.2,]{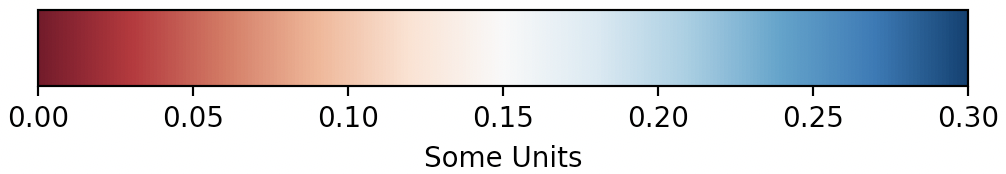}}
\end{tabular}

\caption{Attention patterns of head 11 across layer 1 (first row), 2 (second row), and 8 (third row) during pretraining. Pretrain steps labeled atop the attention map. We averaged the attention maps of different input sentences to get the attention pattern of a single head.}
\label{fig:attention}
\end{figure}
This implies the last few layers of ALBERT learn faster than the top few layers. 
This phenomenon is also revealed by observing the attention patterns across different layers during pretraining.
Figure~\ref{fig:attention} shows that the diagonal attention pattern ~\citep{kovaleva-etal-2019-revealing} of layer 8 emerges earlier than layer 2, with the pattern of layer 1 looms the last\footnote{GIF files are provided in this website: \url{https://albertembryo.wordpress.com/}}.

\begin{figure}[ht]
\begin{subfigure}{0.48\textwidth}
\centering
\includegraphics[clip, trim = 15px 2px 10px
10px,width=\linewidth]{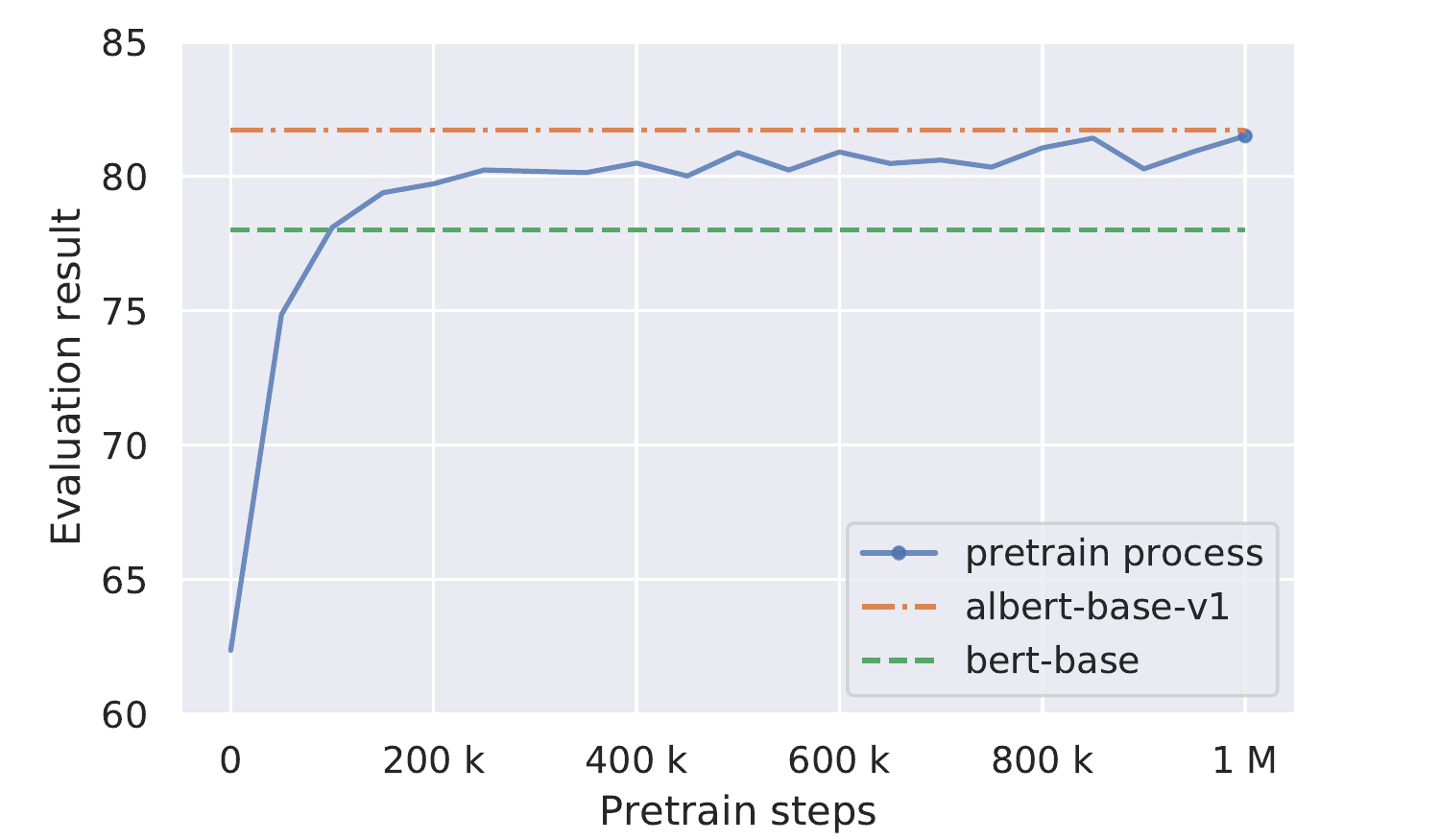}
\caption{GLUE scores over pretraining. GLUE scores of albert-base-v1 and bert-base are also shown by horizontal lines.\footnotemark.}
\label{fig:glue}
\end{subfigure}
\begin{subfigure}{0.48\textwidth}
\centering
\includegraphics[clip, trim = 15px 2px 10px
10px,width=\linewidth]{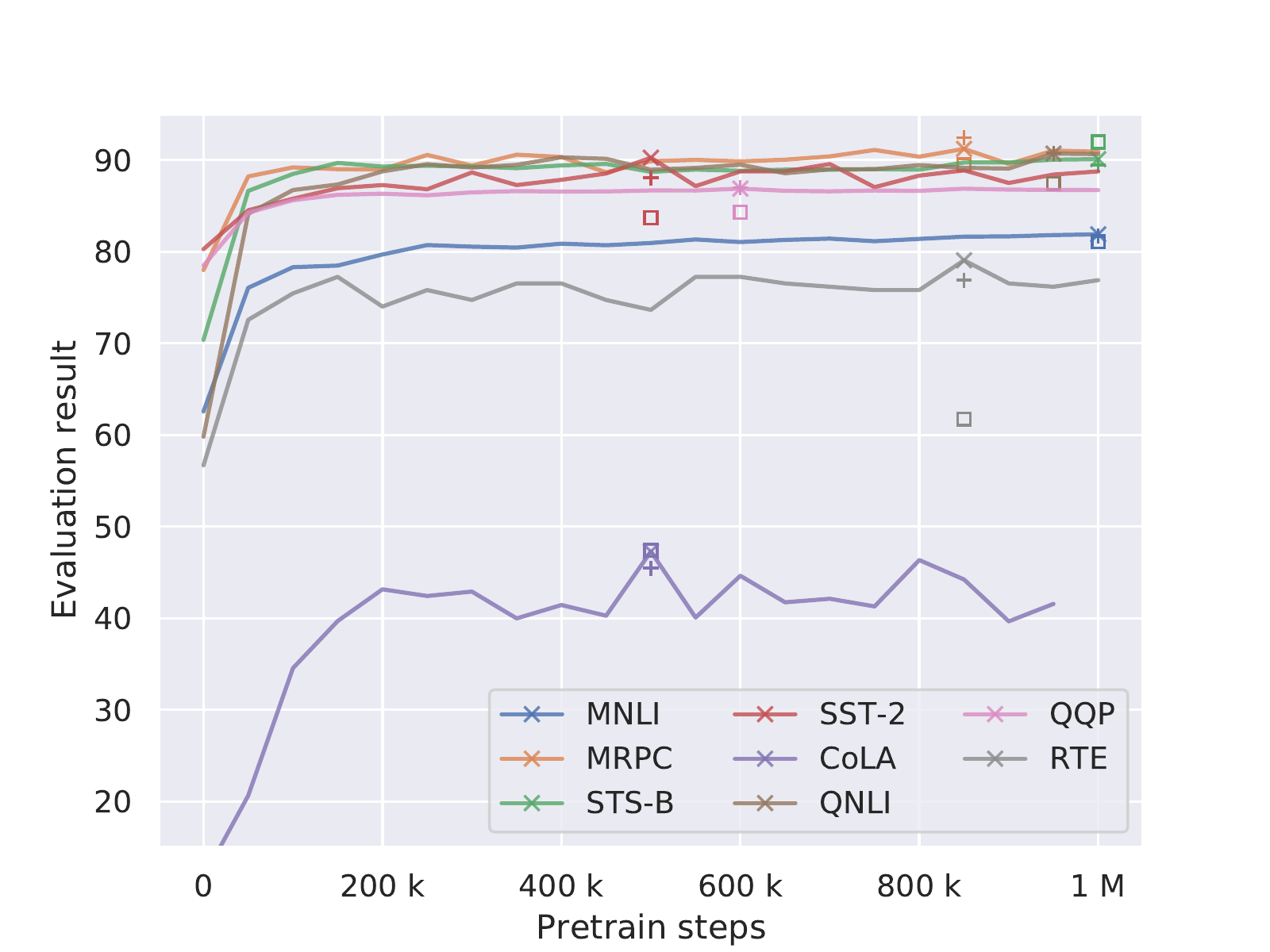}
\caption{Performance of individual tasks in GLUE benchmark. Best result during pretraining marked with `x'.
Performances of albert-base-v1 and bert-base-uncased are marked with `+' and square respectively.}
\label{fig:glue_individual}
\end{subfigure}
\caption{Downtream evaluation of ALBERT on development set every 50k pretrain steps.
GLUE score is averaged among all tasks except WNLI.
Evaluation metrics: MRPC and QQP: F1, STS-B: Spearman corr., others: accuracy.
The result of MNLI is the average of matched and mismatched. 
}
\label{fig:glue_figure}
\end{figure}
\footnotetext{GLUE score of albert-base-v1 and bert-base are obtained by finetuning ALBERT and BERT models from HuggingFace\citep{Wolf2019HuggingFacesTS}}

\section{Does Expensive and Lengthy Pretraining Guarantee Exceptional Results on Downstream Tasks?}
\label{section: downstream}
While \citet{devlin2018bert} and \citet{lan2019albert} have shown that more pretrain steps lead to better GLUE scores, whether the performance gain of downstream tasks is proportional to the resources spent on additional pretrain steps is unknown.
This drives us to explore the downstream performance of the ALBERT model before fully pretrained.  
We choose GLUE benchmark ~\citep{wang2018glue} for downstream evaluation, while excluding WNLI, following ~\citet{devlin2018bert}.

We illustrate our results of the downstream performance of the ALBERT model during pretraining in Figure~\ref{fig:glue_figure}.
While the GLUE score gradually increases as pretraining proceeds, the performance after 250k does not pale in comparison with a fully pretrained model (80.23 v.s. 81.50).
From Figure~\ref{fig:glue_individual}, we also observe that most GLUE tasks reach comparable results with their fully pretrained counterpart over 250k pretrain steps, except for MNLI and QNLI, indicating NLI tasks do benefit from more pretrain steps when the training set is large.

We also finetuned BERT and ELECTRA models as pretraining proceeds, and we observe similar trends.
The GLUE scores of the BERT and ELECTRA model rise drastically in the first 100k pretrain steps, and then the performance increments less slowly afterward.
We put the detailed result of these two models in Section~\ref{app:downstream_bert}.

We conclude that it may not be necessary to train an ALBERT model until its pretraining loss converges to obtain exceptional downstream performance. 
The majority of its capability for downstream tasks has already been learned in the early stage of pretraining.
Note that our results do not contradict previous findings in ~\citet{devlin2018bert}, ~\citet{liu2019roberta}, and ~\citet{clark2019electra}, all of which showing that downstream tasks do benefit from more pretrain steps; we show that the performance gain on downstream tasks in latter pretrain steps might be disproportional to the cost on more pretrain steps.

\section{World Knowledge Development During Pretraining}
\label{section: knowledge}
~\citet{petroni2019language} has reported that language models contain world knowledge. 
To examine the development of world knowledge of a pretrained language model, we conduct the same experiment as in ~\citet{petroni2019language}. We use a subset of T-REx ~\citep{elsahar2018t} from the dataset provided by ~\citet{petroni2019language} to evaluate ALBERT's world knowledge development. 

\begin{figure}[h]
\centering
\includegraphics[clip, trim = 14px 2px 39px 35px,width=0.95\linewidth]{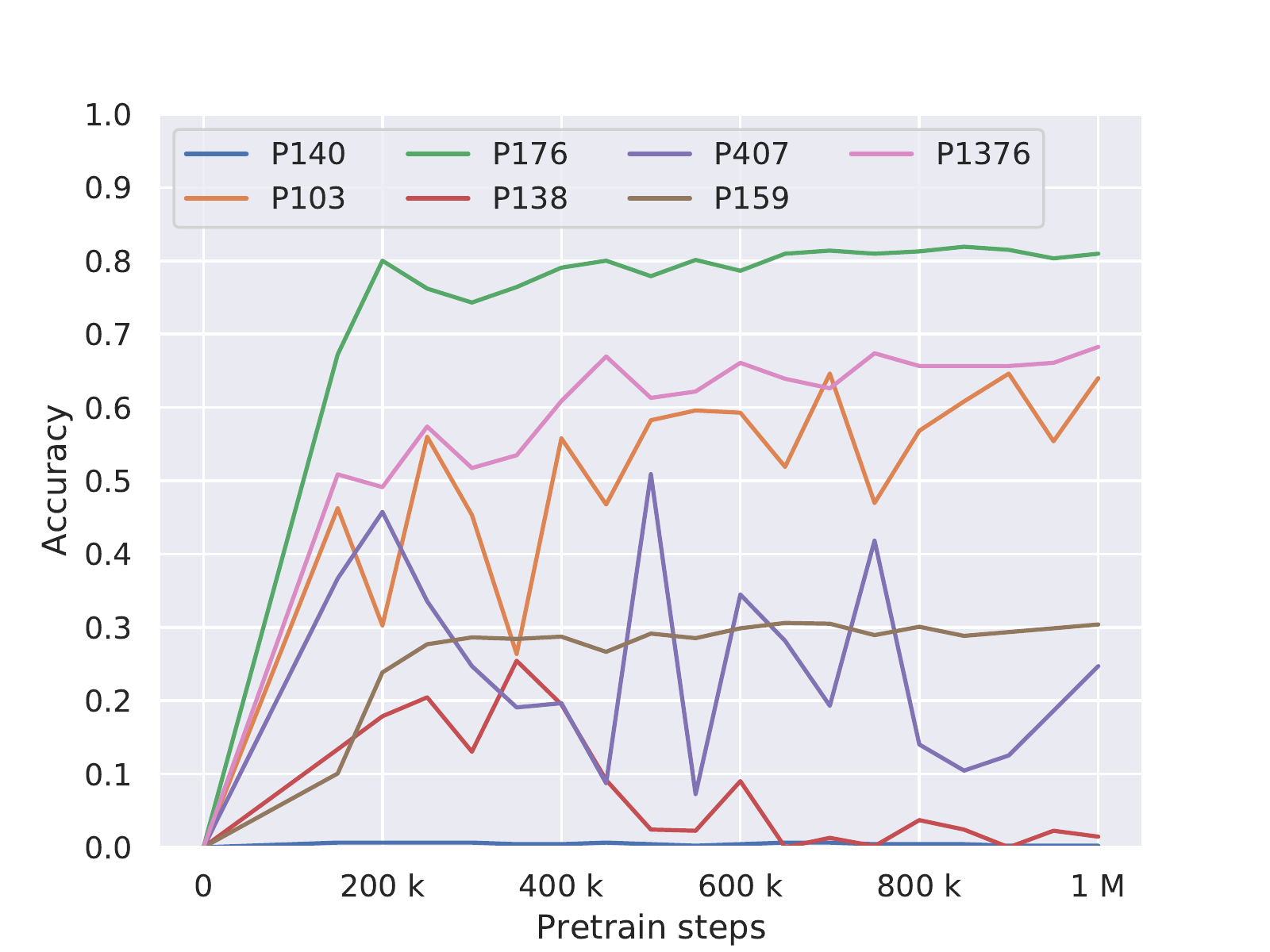}
\caption{World knowledge development during pretraining evaluated every 50k pretrain steps.
Types of relation, and template are shown in Table \ref{tab:rel_table}}
\label{fig:know}
\end{figure}

The results are shown in Figure~\ref{fig:know}, in which we observe that world knowledge is indeed built up during pretraining, while performance fluctuates occasionally.
From Figure~\ref{fig:know}, it is clear that while some types of knowledge stay static during pretraining, some vary drastically over time, and the result of a fully pretrained model (at 1M steps) may not contain the most amount of world knowledge.
We infer that world knowledge of a model depends on the corpus it has seen recently, and it tends to forget some knowledge that it has seen long ago.
These results imply that it may not be sufficient to draw a conclusion on ALBERT's potential as a knowledge base merely based on the final pretrained one's behavior.
We also provide qualitative results in Appendix~\ref{app: qualitity}.

\begin{table}[h]
    \centering
    \begin{tabular}{c|c}
    \centering
        Type & Query template\\
        \hline
        P140 &  [X] is affiliated with the [Y] religion . \\
P103 & The native language of [X] is [Y] . \\
P176 &  [X] is produced by [Y] . \\
P138 & [X] is named after [Y] . \\
P407 &[X] was written in [Y] . \\
P159 & The headquarter of [X] is in [Y] . \\
P1376 &  [X] is the capital of [Y] . \\
    \end{tabular}
    \caption{Relations in Figure~\ref{fig:know}. We fill in [X] with the subject, [Y] with [MASK] and ask model to predict Y.}
    \label{tab:rel_table}
\end{table}
\section{Conclusion}
Although finetuning from pretrained language models puts in phenomenal downstream performance, the reason is not fully uncovered.
This work aims to unveil the mystery of the pretrained language model by looking into how it evolves. 
Our findings show that the learning speeds for reconstructing and predicting tokens differ across POS.
We find that the model acquires semantic and syntactic knowledge simultaneously at the early pretraining stage.
We show that the model is already prepared for finetuning on downstream tasks at its early pretraining stage. 
Our results also reveal that the model's world knowledge does not stay static even when pretraining loss converges.
We hope our work can bring more insights into what makes a pretrained language model a pretrained language model.
\section*{Acknowledgements}
We thank all the reviewers' valuable suggestions and efforts towards improving our manuscript.
This work was supported by Delta Electronics, Inc.
We thank to National Center for High-performance Computing (NCHC) for providing computational and storage resources.
\clearpage
\bibliography{emnlp2020}
\bibliographystyle{acl_natbib}
\clearpage
\newpage
\appendix
\section{Modifications from the Reviewed Version}
We made some modifications in the camera-ready version, mostly based on the reviewers' recommendations and for better reproducibility.
\begin{itemize}
    \item We add the result of BERT and ELECTRA in Section~\ref{section: mask}, Section~\ref{sec:probe}, and Section~\ref{section: downstream}.
    \item We reimplement the source code for Section~\ref{sec:probe} and renew the experiment results accordingly. While the exact values are slightly different, the general trends are the same and do not affect our observation.
    \item We add the results of coreference resolution in our probing experiments, following the reviewers' suggestion.
    \item We polish our wordings and presentations in text and figures.
\end{itemize}
\section{Pretraining}
\subsection{ALBERT}
\label{app: pretrain hpp}
As mentioned in the main text, we only modified a few hyperparameters to fit in out limited computation resources, listed in Table~\ref{tab:pretrain hyperparam}.
The Wikipedia corpus used in our pretraining can be download from \url{https://dumps.wikimedia.org/enwiki/latest/enwiki-latest-pages-articles.xml.bz2}, and the Gutenburg dataset can be download from \url{https://web.eecs.umich.edu/~lahiri/gutenberg_dataset.html}. 
The number of parameters in our ALBERT model is 12M.
\begin{table}[h]
    \centering
    \begin{tabular}{|c|c|}
    \hline
        Batch size & 512 \\
        Learning rate & 6.222539674E-4 \\
        Total steps & 1M \\
        Warmup steps & 25k\\
    \hline
    \end{tabular}
    \caption{Pretraining hyperparemeters for ALBERT.}
    \label{tab:pretrain hyperparam}
\end{table}

\subsection{BERT}
We use the same dataset as we trained ALBERT to pretrain BERT.
We pretrained a BERT-base-uncased model using the official implementation of BERT at \url{https://github.com/google-research/bert}, and we follow all hyperparameters of the original implementation.
Note that the ~\citet{devlin2018bert} mentioned they trained BERT with a maximum sequence length of 128 for the first 900K steps, and then trained the model with a maximum sequence length 512 for the rest 100K steps; we follow this training procedure.
The number of parameters in our BERT model is 110M.

\subsection{ELECTRA}
We use OpenWebTextCorpus~\citep{Gokaslan2019OpenWeb} from \url{https://skylion007.github.io/OpenWebTextCorpus/} to pretrain an Electra-base model.
We pretrained this model using the official implementation of ELECTRA at \url{https://github.com/google-research/electra}, and we follow all hyperparameters of the original implementation.
The number of parameters in our ELECTRA model used for finetuning (the discriminator part) is 110M.

\section{Mask Predict and Token Reconstruction}
\subsection{Dataset}
As mentioned in Section~\ref{section: mask}, we use the POS annotations in OntoNotes 5.0, and we only use the CoNLL-2012 test set for our experiments.
While there are 48 POS labels, we only report the mask prediction and token reconstruction of a much smaller subset---those we are more familiar with.
The statistics of these POS are in Table~\ref{tab:pos_count}.
\begin{table}[h]
    \centering
    \begin{tabular}{c|c}
        POS & Count \\
        \hline
        Conjunction & 5109\\
        Determiner & 14763\\
        Preposition & 18059\\
        Adjective & 9710\\
        Adverb & 7992\\
        Verb (all forms) & 21405\\
        Noun & 29544\\
        Proper noun & 13144\\
    \end{tabular}
    \caption{Statistics of POS used in experiments.}
    \begin{tabular}{c|c}
        Verb form & Count \\
        \hline
        Base form &  5865\\
        Past tense & 5398 \\
        Gerund or present participle & 2821\\
        Past participle & 3388\\
        3rd person singular present&3933\\
        
    \end{tabular}
    \label{tab:pos_count}
    \caption{Statistics of different verb forms used in experiments.}
\end{table}
\subsection{Mask Predict and Token Reconstruction of Different Verb Forms}
\label{app: verb}
\begin{figure}[h]
\centering
\begin{subfigure}{0.45\textwidth}
\includegraphics[width=\linewidth]{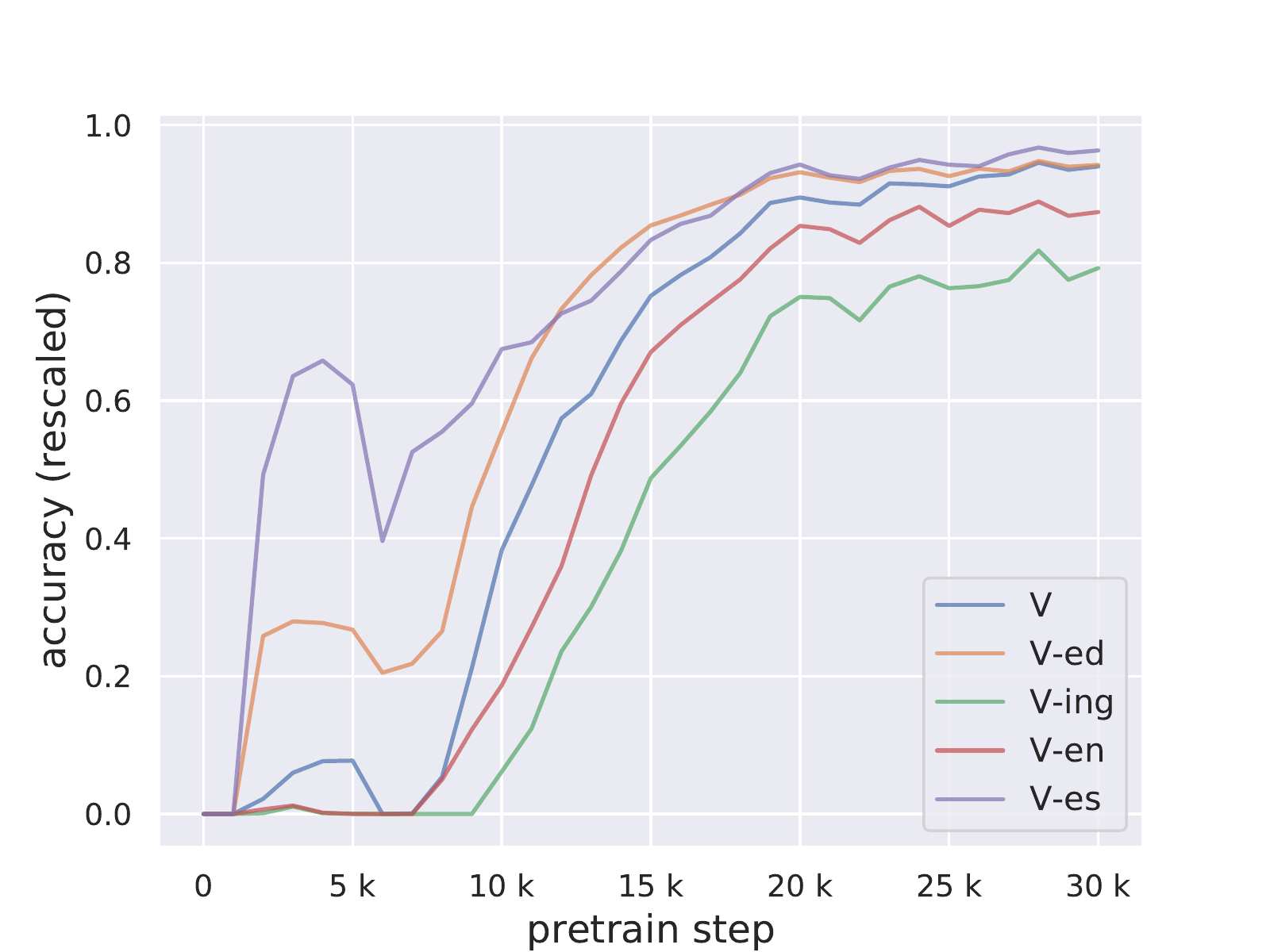}
\caption{Token reconstruction.}
\label{fig:recon_verb}
\end{subfigure}
\begin{subfigure}{0.45\textwidth}
\includegraphics[width=\linewidth]{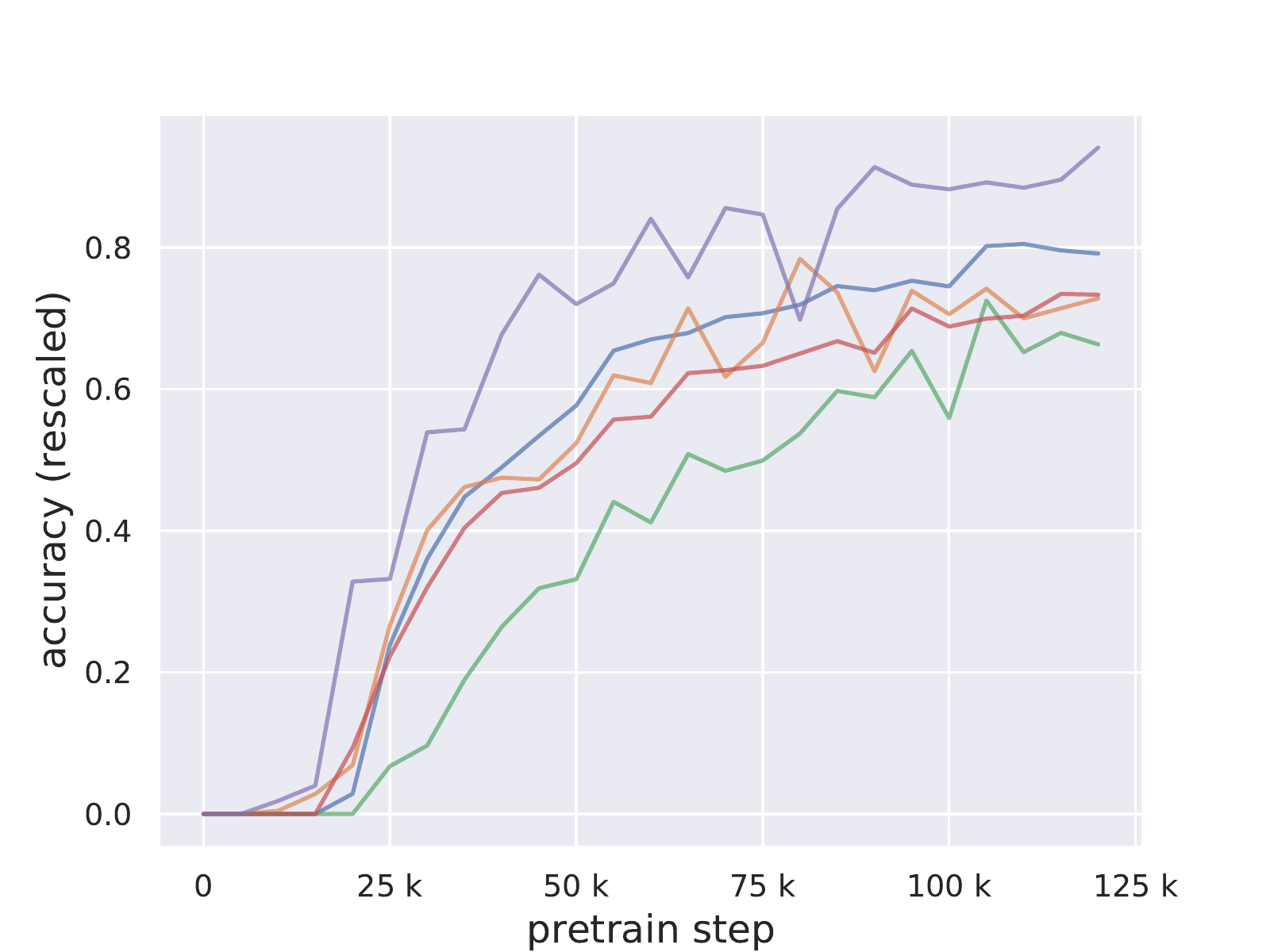}
\caption{Mask prediction.}
\label{fig:mask_pred_verb}
\end{subfigure}
\caption{Token reconstruction (\ref{fig:recon_verb}) and mask prediction (\ref{fig:mask_pred_verb}) accuracy.  We also rescale the accuracy as in Figure~\ref{fig:reconstruction_and_mask_predict}.}
\label{fig:reconstruction_and_mask_predict_verb}
\end{figure}

We provide supplementary materials for Section~\ref{section: mask}.
In Figure~\ref{fig:reconstruction_and_mask_predict_verb}, we observe that ALBERT learns to reconstruct and predict verb of different forms at different times.
The average occurrence rate of verb in different form from high to low is V-es, V-ed, V, V-en, V-ing, which coincides with the priority being leaned.

\subsection{How Does Occurrence Frequency Affect Learning Speed of A Word?}
\begin{figure}[t]
\centering
\includegraphics[clip, trim = 10px 2px 39px
40px,width=\linewidth]{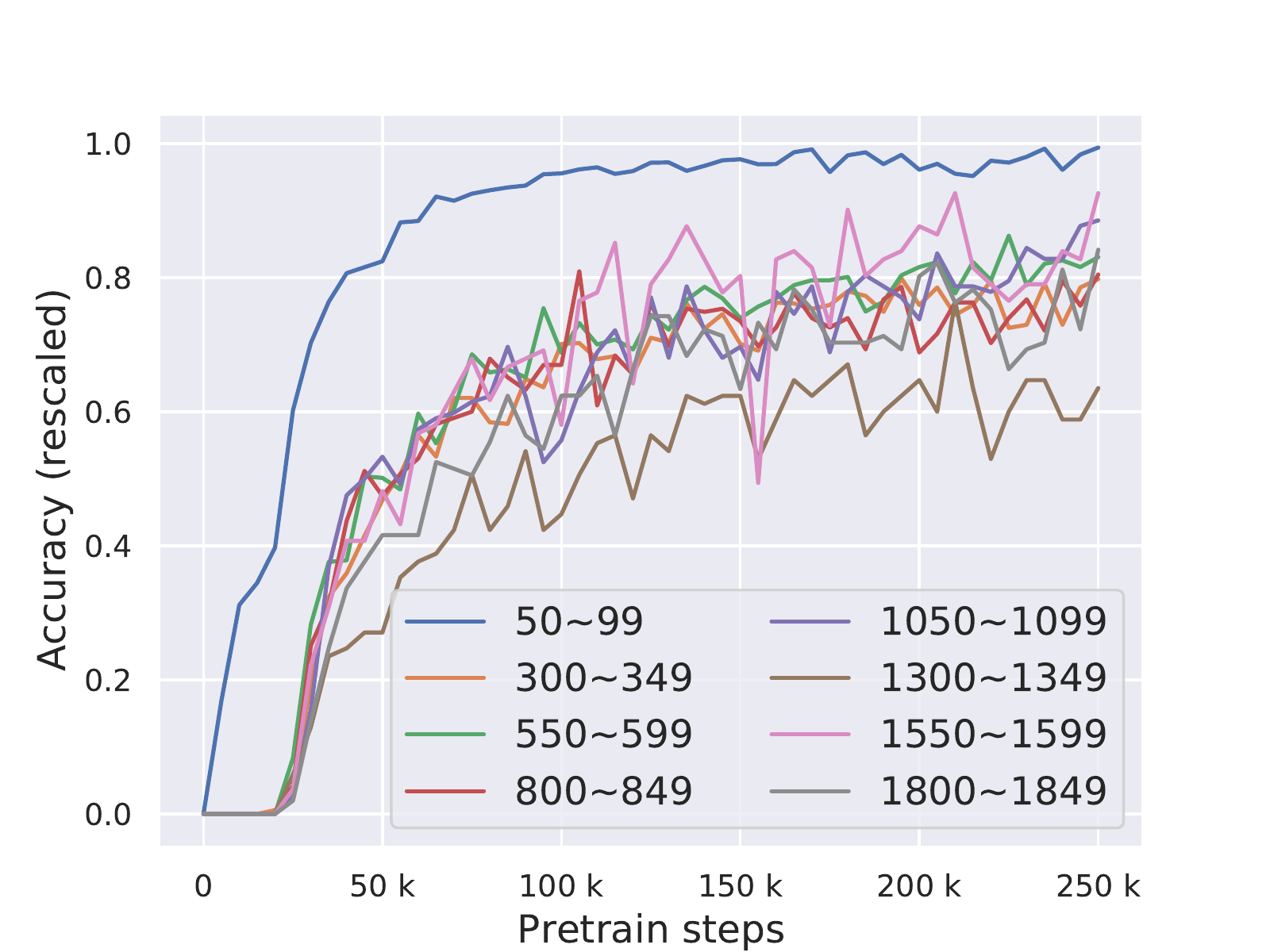}
\caption{Rescaled mask prediction accuracy for different frequency. 50{\raise.17ex\hbox{$\scriptstyle\sim$}}99 means the top 50 to top 99 occurring tokens}
\label{fig:frequency}
\end{figure}

In the main text, we observe that words of different POS are learned at different times of pretraining. 
We also pointed out that the learning speed of different POS roughly corresponds to their occurrence rate.
However, it is not clear to what extent a word's occurrence frequency affects how soon it can be learned to reconstruct or mask-predict by the model.
We provide a deeper analysis of the relationship between the learning speed of a word and its occurrence rate in Figure~\ref{fig:frequency}.
We observe from Figure~\ref{fig:frequency} that the top 50 to 99 occurring tokens are indeed learned faster than other words which occur lesser.
However, as for the top 300 to 349 occurring tokens and the top 1550 to 1599 occurring tokens, it is unclear which ones are learned earlier.
We can infer from Figure~\ref{fig:frequency} and Figure~\ref{fig:mask_pred} that the occurring rate and POS of a word both contribute to how soon the model can learn it to some extent. 

\begin{table*}[ht]
    \centering
    \begin{tabular}{ccccc}
        \hline
    Task & \(|\mathcal{L}|\) & Examples & Tokens & Total Targets\\
    \hline
    POS & 48 & 116K / 16K / 12K&  2.2M / 305K / 230K & 2.1M / 290K / 212K \\
    Constituent &  30 & 116K / 16K / 12K & 2.2M / 305K / 230K & 1.9M / 255K / 191K\\
    SRL &  66 & 253K / 35K / 24K & 6.6M / 934K / 640K & 599K / 83K / 56K\\
    \hline
    \end{tabular}
    \caption{Statistics of the number of labels, examples, tokens
and targets (split by train/dev/test) we used in probing experiments. \(| \mathcal{L}|\) denotes number of target labels.}
    \label{tab: onto}
\end{table*}
\subsection{Mask Predict and Token Reconstruction of BERT}
\label{app:mask_predict_bert}
\begin{figure}[h]
\centering
\begin{subfigure}{0.45\textwidth}
\includegraphics[width=\linewidth]{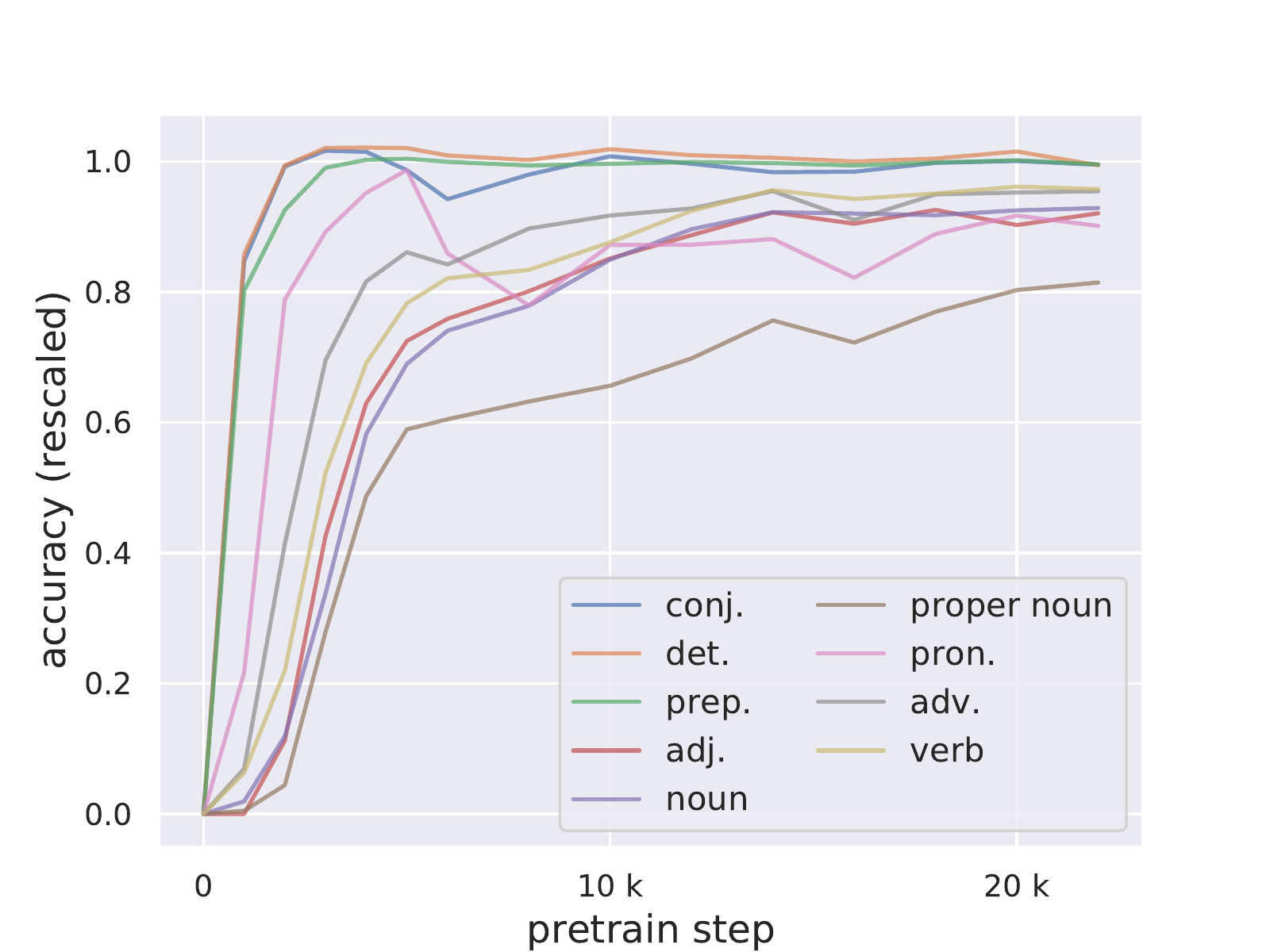}
\caption{Token reconstruction of BERT}
\end{subfigure}
\begin{subfigure}{0.45\textwidth}
\includegraphics[width=\linewidth]{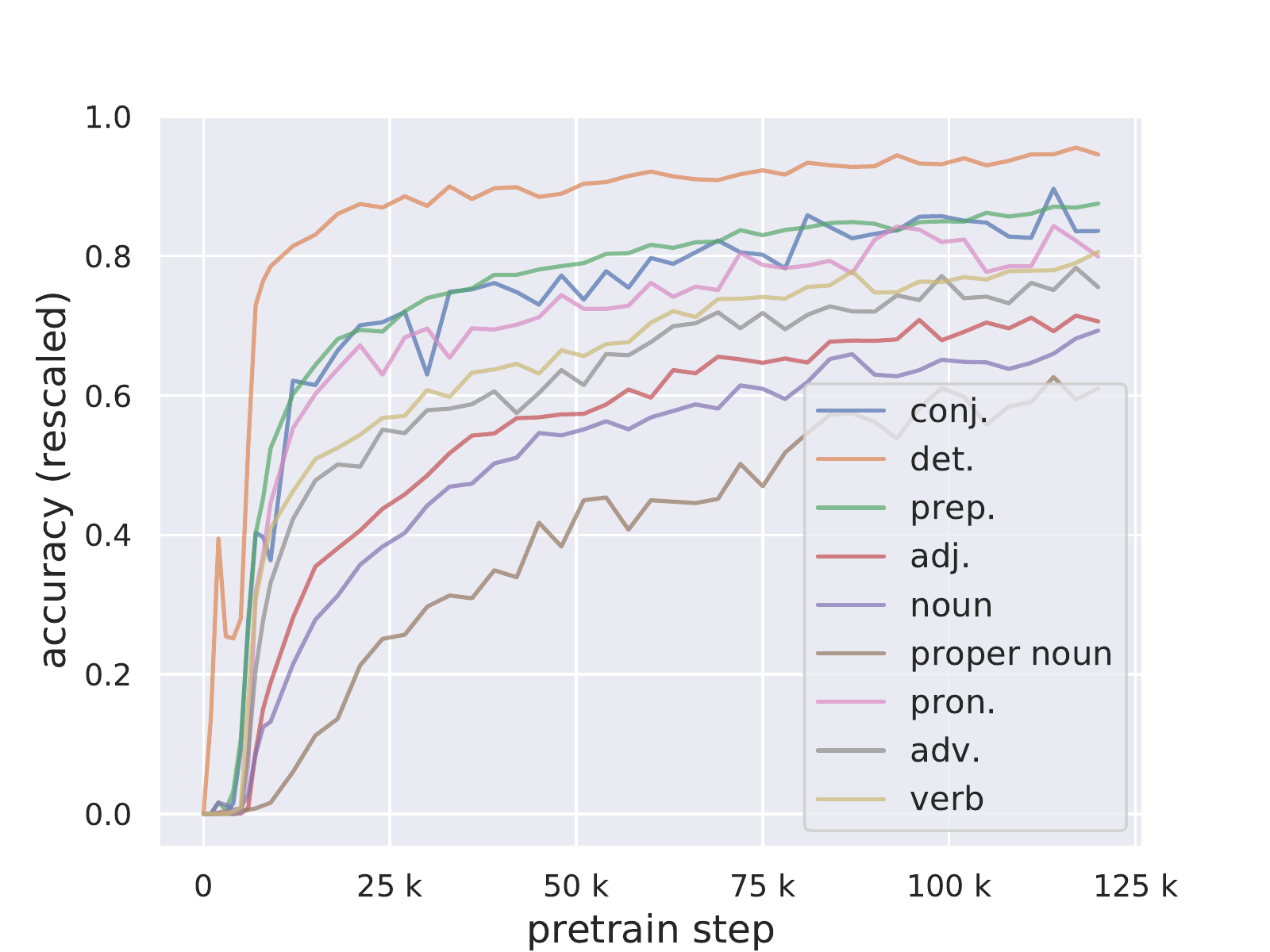}
\caption{Mask prediction of BERT}
\end{subfigure}
\caption{ We also rescale the accuracy as in Figure~\ref{fig:mask_pred}.}
\label{fig:reconstruction_and_mask_predict_bert}
\end{figure}

We provide the results of BERT's token reconstruction and mask prediction in Figure~\ref{fig:reconstruction_and_mask_predict_bert}.
We observe content words are learned later than function words, while the learning speed is faster than ALBERT.
To be more specific, we say a word type \(A\) is learned faster than another word type \(B\) if either the learning curve of \(A\) rises earlier than \(B\) from 0, or if the rescaled learning curve of \(A\) is steeper than that of \(B\).

\section{Probing Experiments}
\subsection{Probing Model Details}
As mentioned in the main text, we modified and reimplemented the edge probing~\citep{tenney2019you} models in our experiments.
The modifications are detailed as follow:
\begin{itemize}
\item We remove the projection layer that projects representation output from the language model to the probing model's input dimension.
\item We use average pooling to obtain span representation, instead of self-attention pooling.
\item We use linear classifiers instead of 2-layer MLP classifiers.
\item We probe the representation of a single layer, instead of concatenating or scalar-mixing representations across all layers.
\end{itemize}
Since our probing models are much simpler than those in \citet{tenney2019you}, probing results might be inferior to the original work.
The number of model's parameters in our experiments is approximately 38K for POS tagging, 24K for constituent tagging, and 100K for SRL.
\subsection{Dataset}
We use OntoNotes-5.0, which can be download from \url{https://catalog.ldc.upenn.edu/LDC2013T19}.
The statistics of this dataset is in Table~\ref{tab: onto}.

\subsection{SRL, Coreference Resolution, and Constituent Labeling Results}
\label{app: probe}
\begin{figure}[h]
\centering
\begin{subfigure}{0.45\textwidth}
\includegraphics[width=0.9\linewidth]{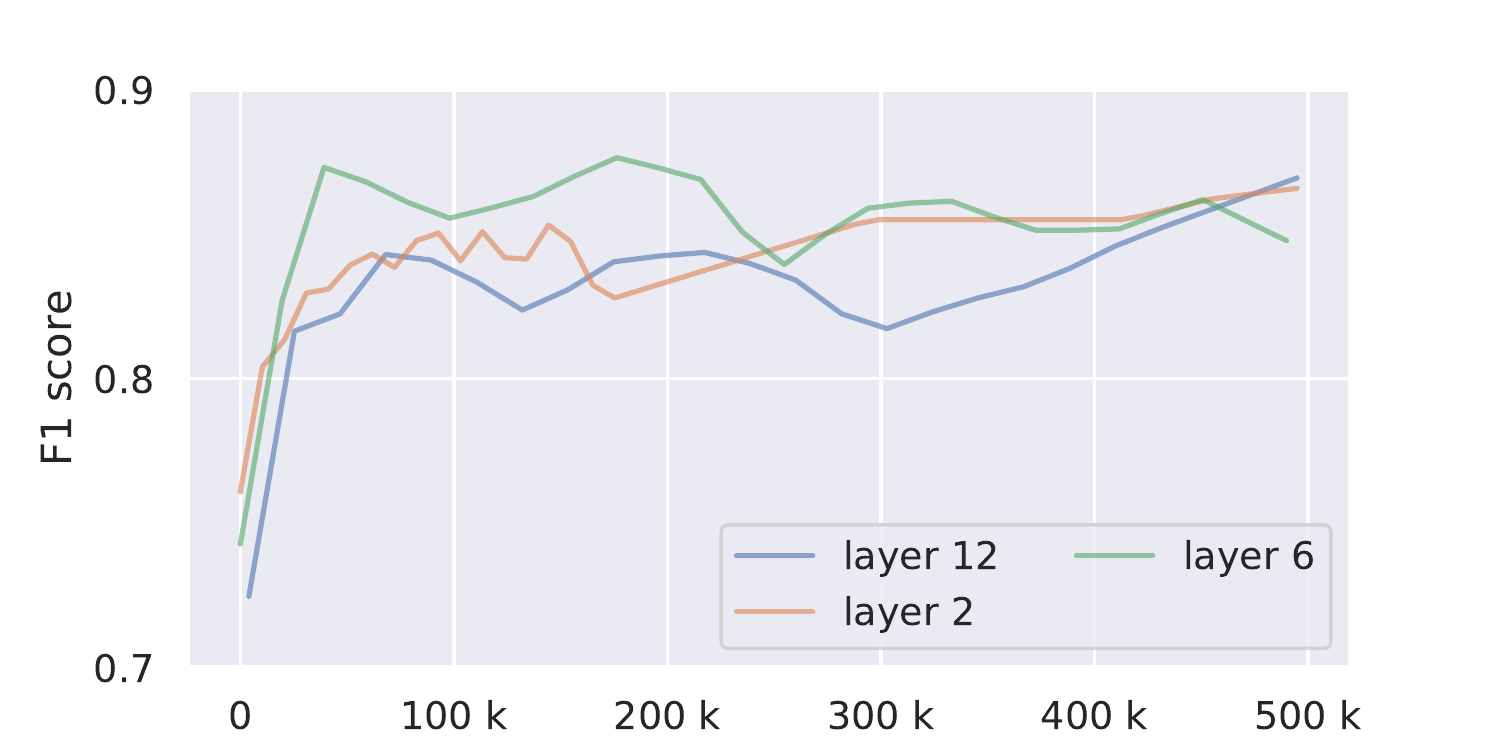}
\caption{Semantic role labeling}
\label{fig:srl}
\end{subfigure}
\begin{subfigure}{0.45\textwidth}
\includegraphics[width=0.9\linewidth]{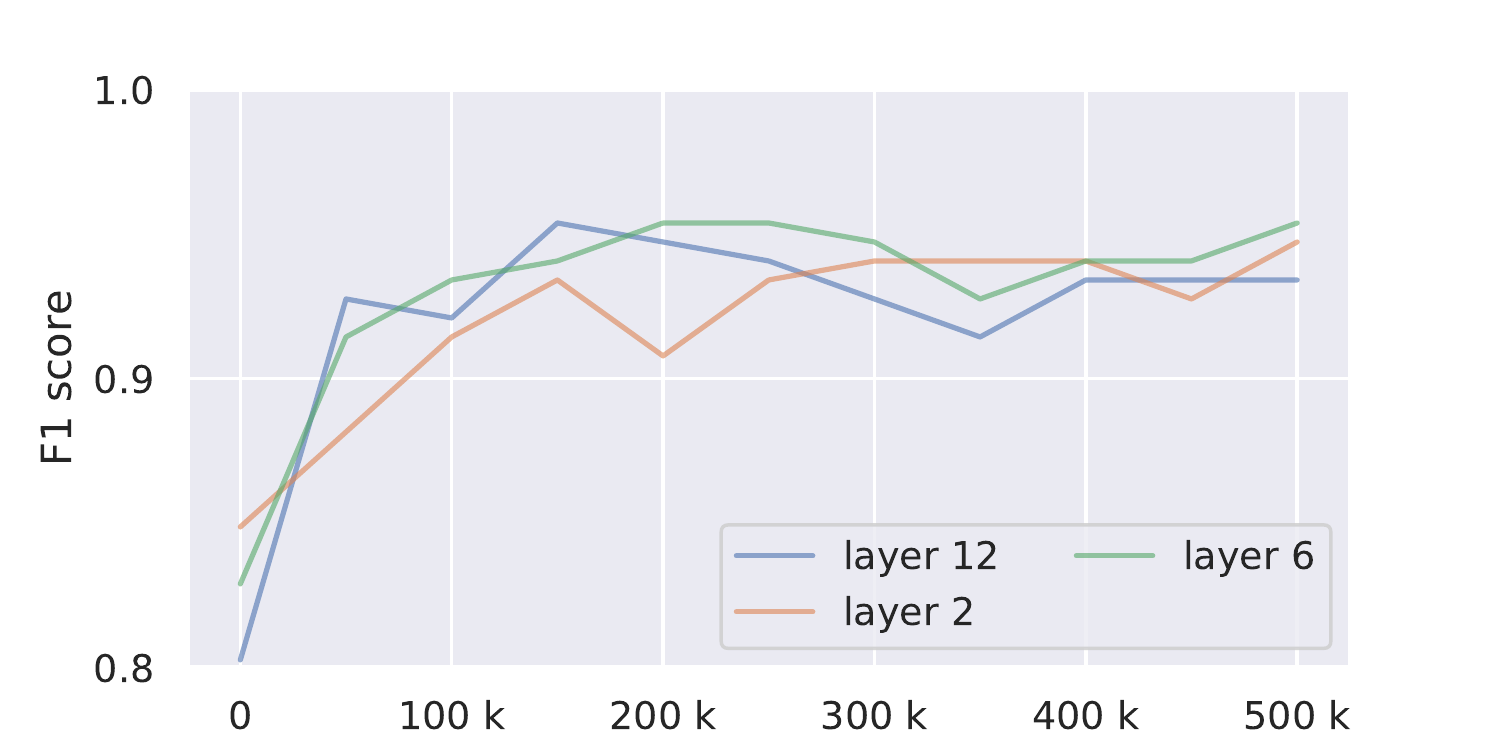}
\caption{Coreference resolution}
\label{fig:coref}
\end{subfigure}
\begin{subfigure}{0.45\textwidth}
\includegraphics[width=0.9\linewidth]{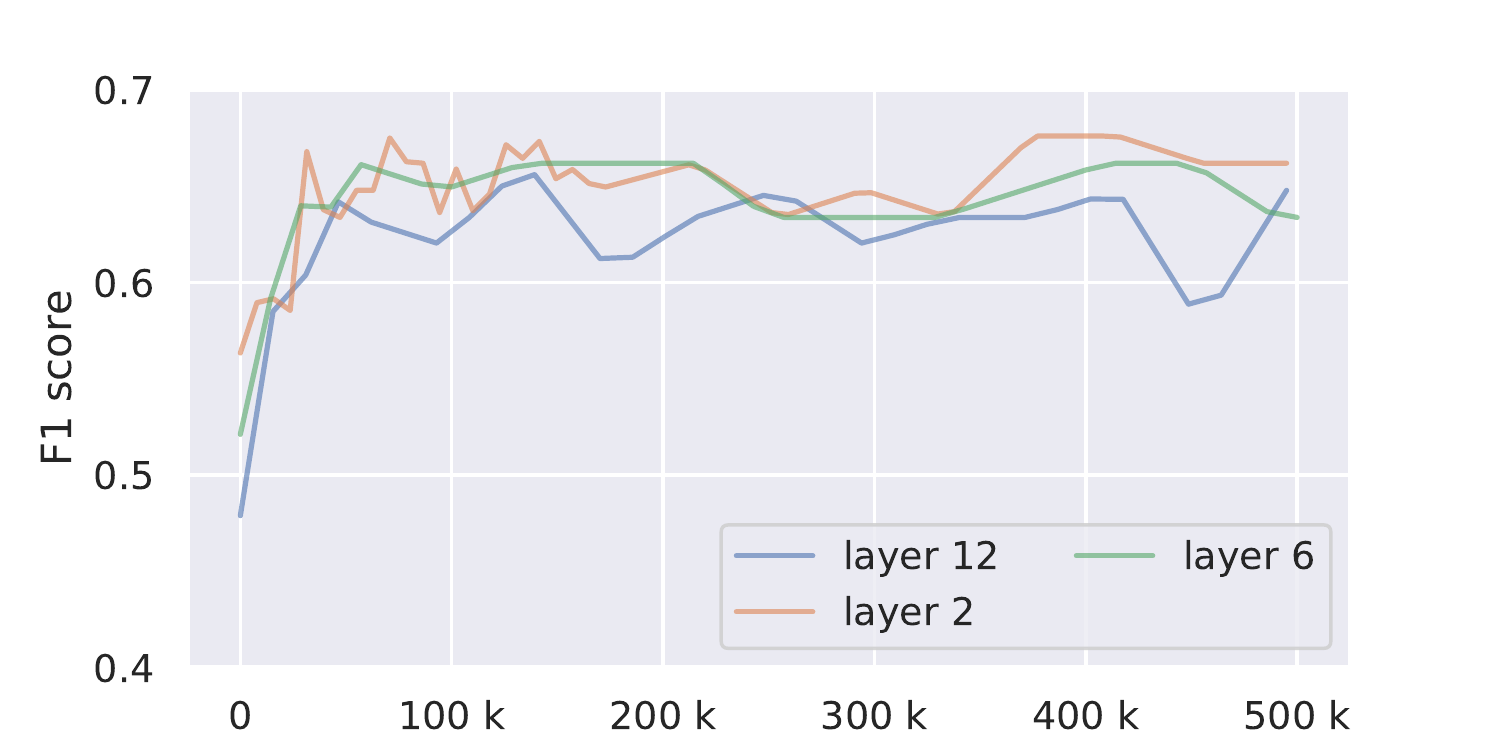}
\caption{Constituent tagging}
\label{fig:const}
\end{subfigure}
\caption{The probing results of SRL (\ref{fig:srl}, coreference resolution (\ref{fig:coref}) and constituency tagging (\ref{fig:const}) during pretraining .
Layers are indexed from the input layer to the output layer, so layer 2 is the output representation from layer 2 of ALBERT. Layers are indexed from 1 to 12.}
\label{fig:probe}
\end{figure}

Here in Figure~\ref{fig:probe}, we show supplementary figures for SRL, coreference resolution, and constituent tagging over 3 of 12 layers in ALBERT for the first 500K pretrain steps.
Together with Figure~\ref{fig:pos}, all four tasks show similar trends.

\subsection{Probing Results of BERT and ELECTRA}
\begin{figure}[h]
\centering
\begin{subfigure}{0.45\textwidth}
\includegraphics[clip, trim = 15px 2px 39px
30px,width=\linewidth]{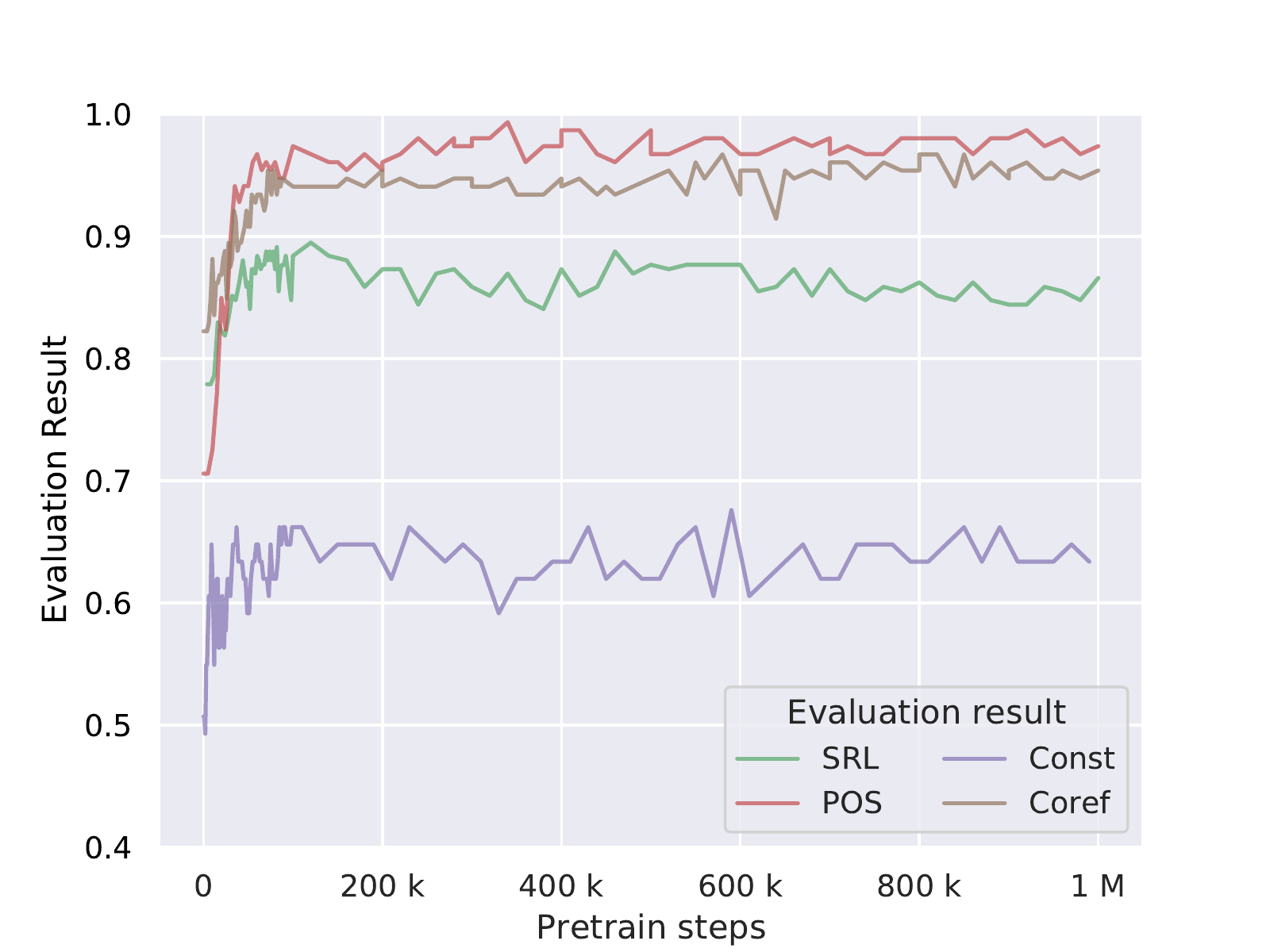}
\caption{Probing results of ALBERT-base model}
\label{fig:bert_probing}
\end{subfigure}
\begin{subfigure}{0.45\textwidth}
\includegraphics[clip, trim = 15px 2px 39px
30px,width=\linewidth]{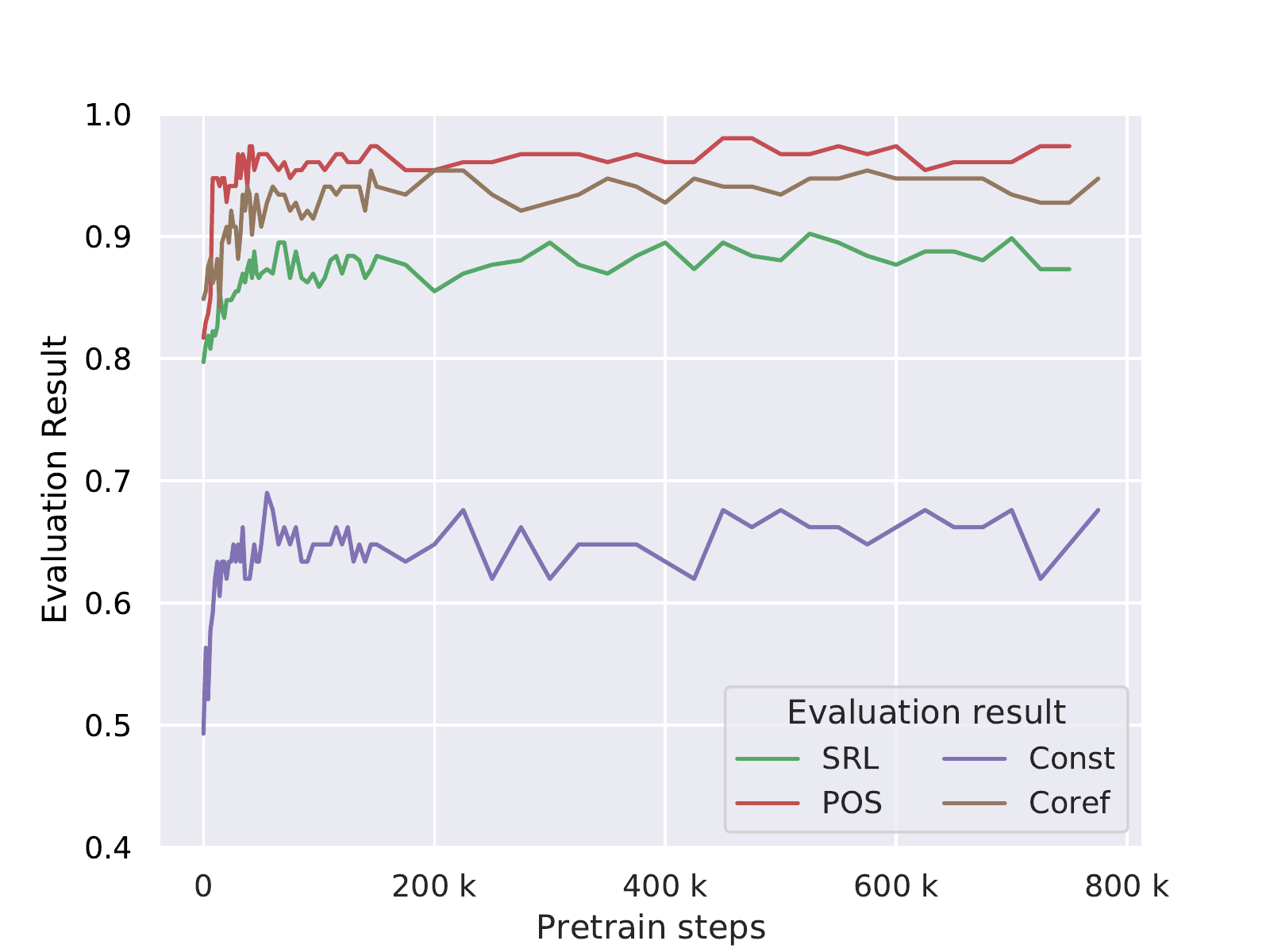}
\caption{Probing results of BERT-base uncased model}
\label{fig:bert_probing}
\end{subfigure}
\begin{subfigure}{0.45\textwidth}
\includegraphics[clip, trim = 15px 2px 39px
30px,width=\linewidth]{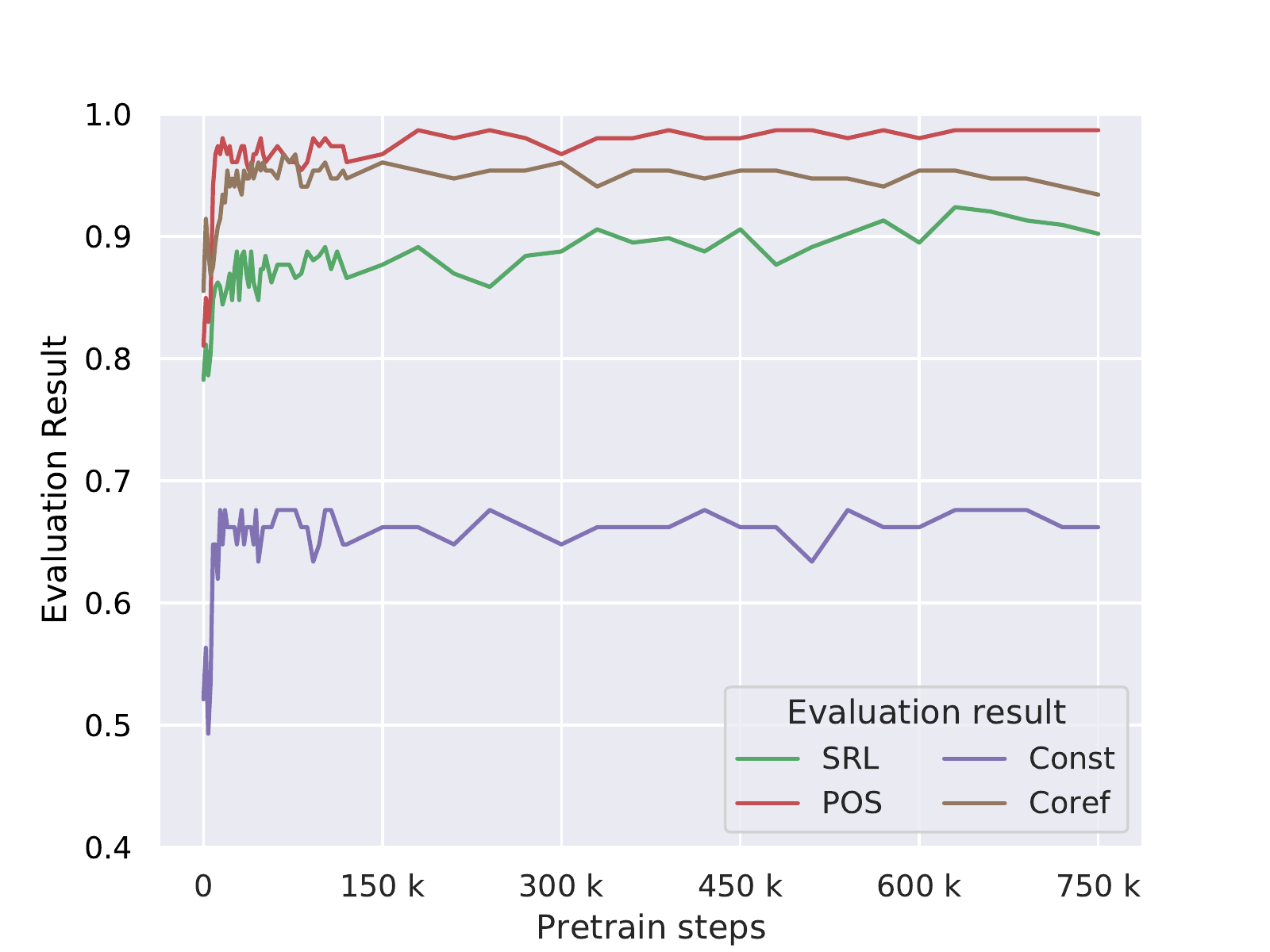}
\caption{Probing results of ELECTRA-base model}
\label{fig:electra_probing}
\end{subfigure}
\caption{Probing results of POS tagging, constituent tagging, semantic role labeling, and coreference resolution, evaluated by micro F1 score.}
\label{fig:probe_bert_electra}
\end{figure}

We provide the probing results of BERT and ELECTRA in Figure~\ref{fig:probe_bert_electra}.
All the probing experiments of ALBERT, BERT, and ELECTRA share the same set of hyperparameters and model architectures.
We observe a similar trend as ALBERT: the probing performance rises quite quickly and plateaus (or even slightly decay) afterward.
We also found that performance drop of those layers closer to ELECTRA's output layers are highly observable, which may spring from its discriminative pretraining nature. 

\section{Downstream Evaluation}
\subsection{Dataset Details}
We provide detail statistics of downstream tasks' dataset in Table~\ref{tab:downstream_st}.
We download GLUE dataset using \url{https://gist.github.com/W4ngatang/60c2bdb54d156a41194446737ce03e2e}, and download SQuAD2.0 dataset from \url{https://rajpurkar.github.io/SQuAD-explorer/}.
\begin{table}[]
    \centering
    \begin{tabular}{c|c}
       Task  & Examples \\
       \hline
       MRPC & 3.6K / 0.4K / 1.7K \\
       RTE & 2.4K / 0.2K / 3K \\
       STS-B & 5.7K / 1.5K / 1.3K \\
       QNLI & 104K / 5.4K / 5.4K \\
       QQP & 363K / 40.4K / 391.0K\\
       CoLA & 8.5K / 1.0K / 1.1K \\
       MNLI & 392.7K / 9.8K + 9.8K / 9.8K + 9.8K\\
       SST-2 & 67.4K / 0.9K / 1.8K\\
       SQuAD2.0 & 13.3K / 11.9K / 8.9K
    \end{tabular}
    \caption{Statistics of (train / dev/ test) in GLUE tasks and SQuAD2.0. 
    MNLI contains matched and mismatched in dev and test set. We didn't evaluate our models' performance on test set.}
    \label{tab:downstream_st}
\end{table}

\begin{figure}[ht!]
\includegraphics[clip, trim = 15px 2px 10px
10px,width=\linewidth]{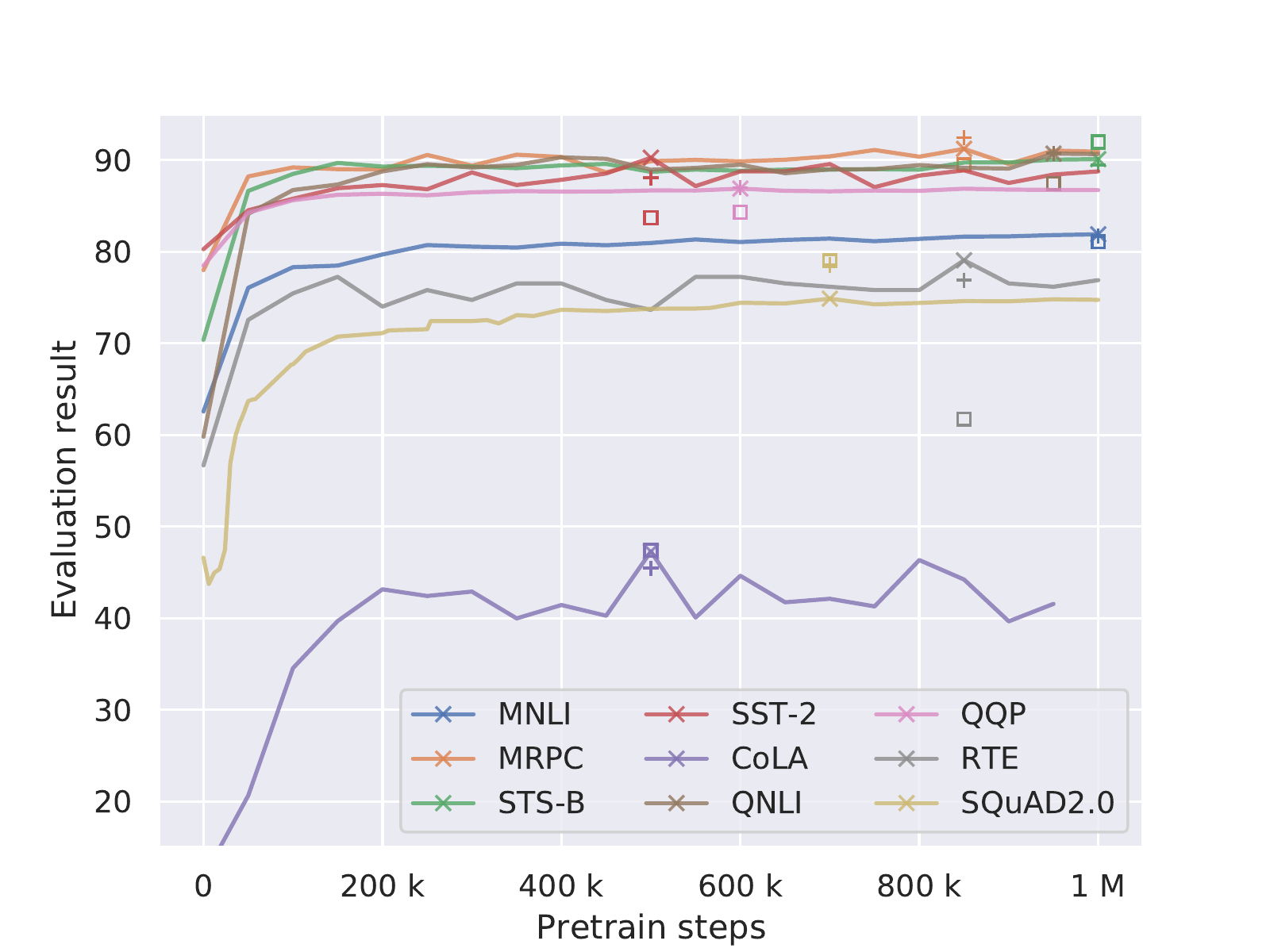}
\caption{Performance of individual tasks in GLUE benchmark, along with SQuAD2.0 result. Best result durining pretraining marked with `x'.
Evaluation metrics: MRPC and QQP: F1, STS-B: Spearman corr., others: accuracy.
The result of MNLI is the average of matched and mismatched.
The result of SQuAD2.0 is the average of F1 and EM scores.
Performances of albert-base-v1 and bert-base-uncased are marked with `+' and square, respectively.}
\label{fig:downstream}
\centering
\end{figure}

\begin{figure}[hb!]
\centering
\begin{subfigure}{0.45\textwidth}
\includegraphics[width=0.9\linewidth]{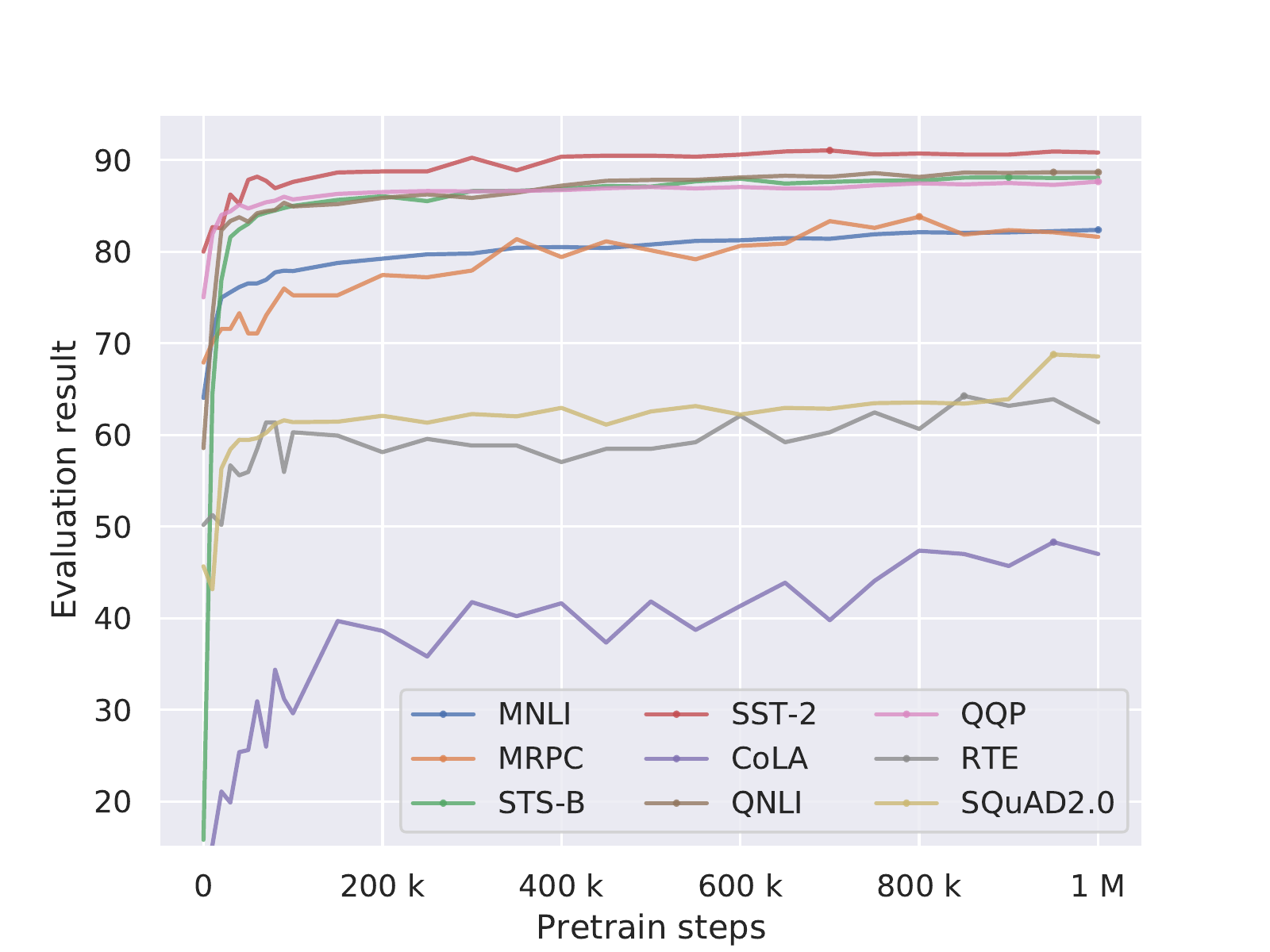}
\caption{GLUE and SQuAD2.0 performances of BERT}
\label{fig:bert_finetune}
\end{subfigure}
\begin{subfigure}{0.45\textwidth}
\includegraphics[width=0.9\linewidth]{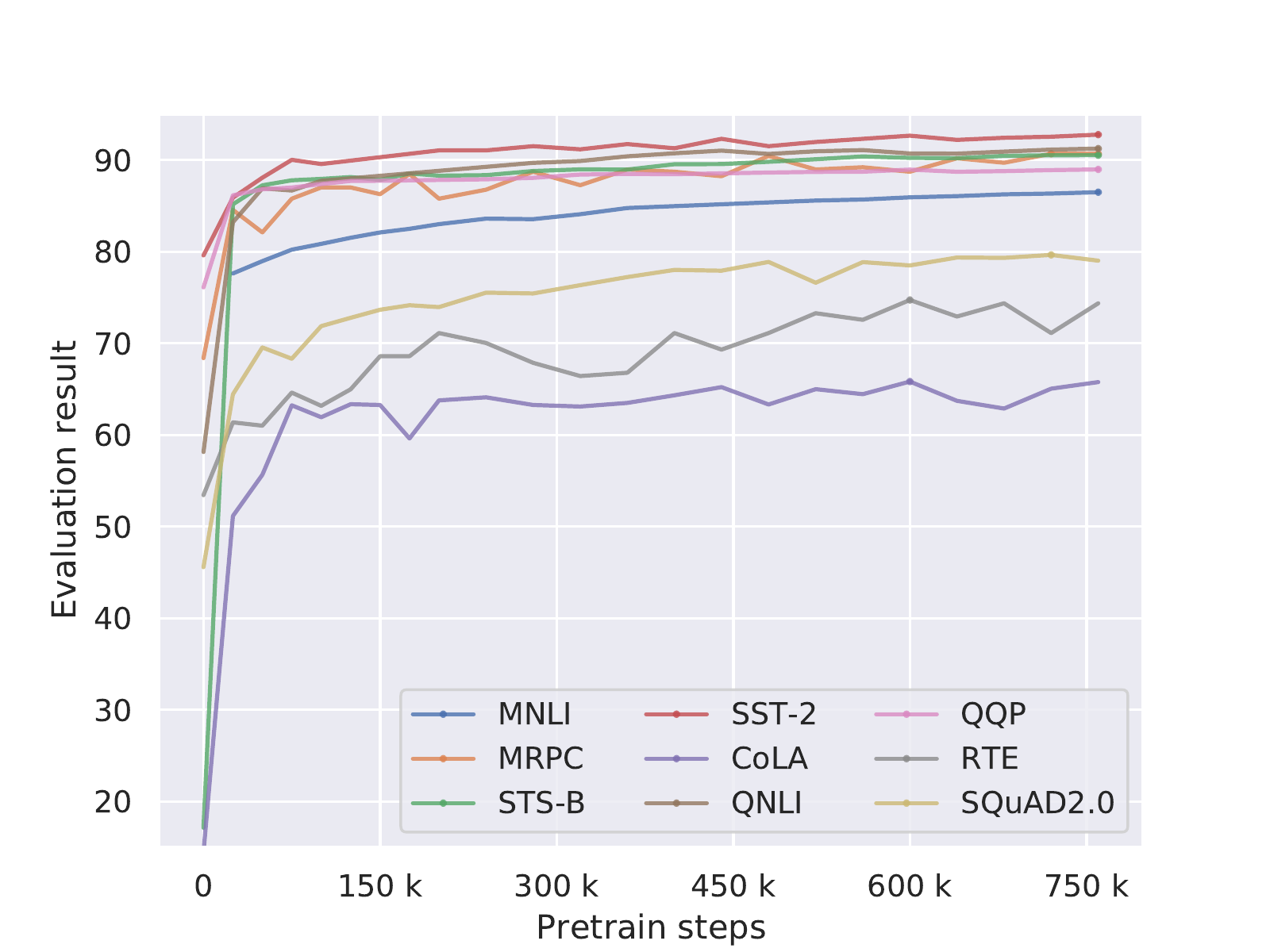}
\caption{GLUE and SQuAD2.0 performances of ELECTRA}
\label{fig:electra_finetune}
\end{subfigure}
\begin{subfigure}{0.45\textwidth}
\includegraphics[width=0.9\linewidth]{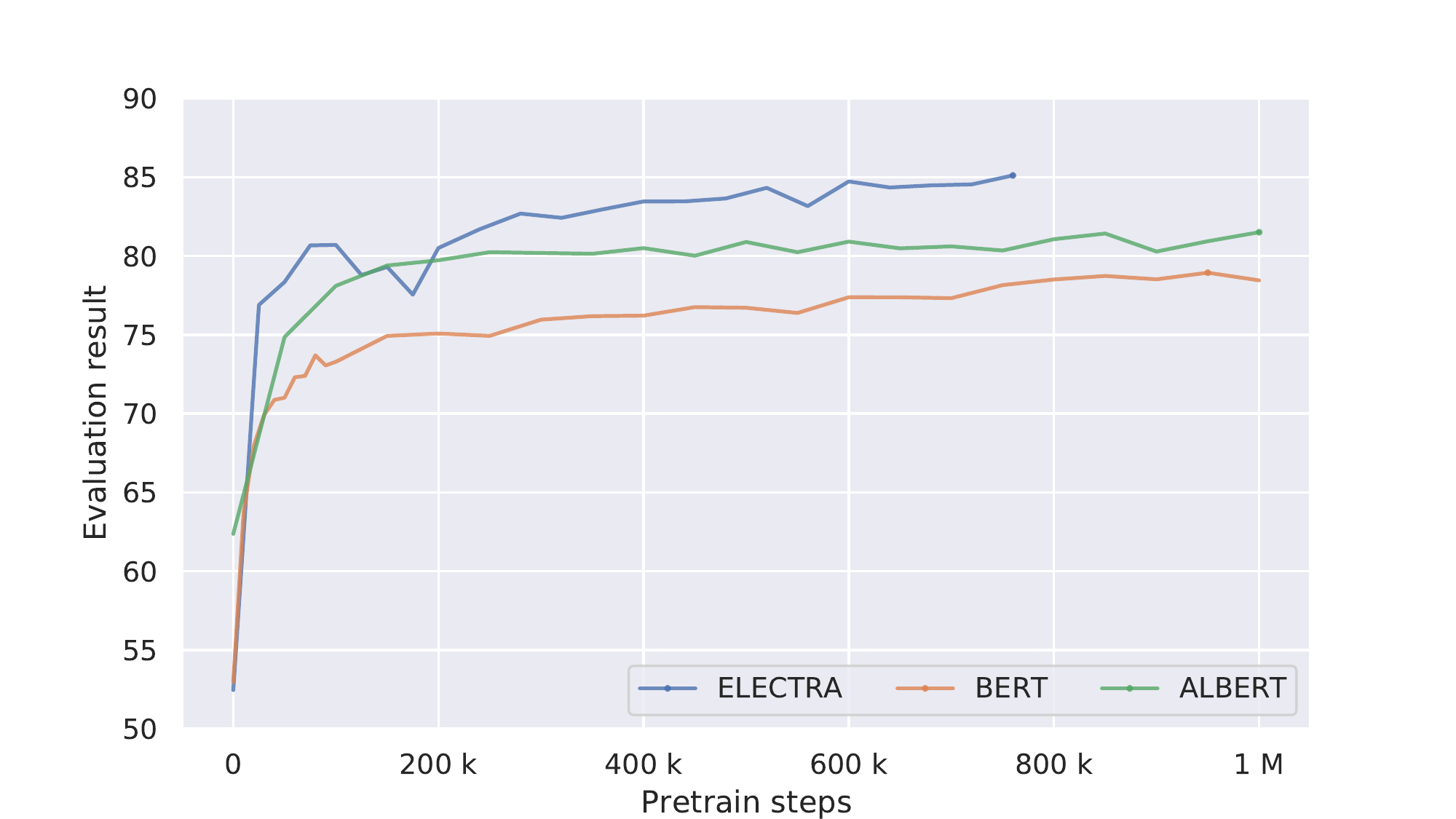}
\caption{GLUE scores of all three models}
\label{fig:glue_all}
\end{subfigure}
\caption{Performance  of  individual  tasks  in  GLUE benchmark, along with SQuAD2.0 result.  Best result durining  pretraining  marked  with  circle.   Evaluation metrics: MRPC and QQP: F1, STS-B: Spearman corr.,others:  accuracy.   The result of MNLI is the averageof matched and mismatched.  The result of SQuAD2.0is the average of F1 and EM scores.}
\label{fig:finetune_others}
\end{figure}

\begin{table*}[ht]
    \centering
    \begin{tabular}{c|cccccccc}
    & LR & BSZ & ALBERT DR & Classifier DR & TS & WS & MSL \\
    \hline
    CoLA &1.00E-05& 16 &0 &0.1&5336 &320 &512 \\
    STS-B &2.00E-05& 16 &0 &0.1 &3598& 214 &512\\
    SST-2& 1.00E-05 &32& 0& 0.1 &20935 &1256 &512 \\
MNLI& 3.00E-05 &128& 0 &0.1 &10000 &1000& 512\\
QNLI& 1.00E-05 &32& 0 &0.1& 33112& 1986& 512\\
QQP& 5.00E-05& 128& 0 &0.1& 14000 &1000 &512\\
RTE &3.00E-05& 32& 0 &0.1 &800 &200 &512\\
MRPC& 2.00E-05& 32& 0 &0.1& 800& 200 &512\\
SQuAD2.0 &3.00E-05& 48& 0 &0.1& 8144& 814 &512
    \end{tabular}
    \caption{Hyperparameters for ALBERT in downstream tasks. LR: Learning Rate. BSZ: Batch Size. DR: Dropout Rate. TS: Training Steps. WS: Warmup Steps. MSL: Maximum Sequence
Length}
    \label{tab:glue_hpp}
\end{table*}

\subsection{Finetune Details}
\label{subsec:finetune}
We use the code in \url{https://github.com/huggingface/transformers/tree/master/examples/text-classification} to run GLUE and use \url{https://github.com/huggingface/transformers/tree/master/examples/question-answering} to run SQuAD2.0.
We provide detailed hyperparameters when we run GLUE benchmark and SQuAD2.0 in Table~\ref{tab:glue_hpp}.
We follow \citet{liu2019roberta} and \citet{lan2019albert}, finetuning RTE, STS-B, and MRPC using an MNLI checkpoint when finetuning ALBERT.
The number of parameters of all downstream tasks is close to the original ALBERT model, which is 12M.

\subsection{Downstream results of ALBERT (with SQuAD2.0)}
\label{app:downstream}
Here we provide performance of individual tasks in GLUE benchmark on development set in Figure~\ref{fig:downstream}, along with performance of SQuAD2.0~\citep{rajpurkar2018know}.

\subsection{Downstream performance of BERT and ELECTRA}
\label{app:downstream_bert}
We use the same hyperparamters in Table\ref{tab:glue_hpp} to finetune BERT and ELECTRA models.
Except for the performance of BERT on SQuAD2.0, all the other results are comparable with those results finetuned from the official Google pretrained models. 
We can observe from Figure~\ref{fig:finetune_others} and Figure~\ref{fig:downstream} that all three models' performance on downstream tasks show similar trends: Performance skyrocketed during the initial pretraining stages, and the return gradually decays later.
From Figure~\ref{fig:glue_all}, we also find that among the three models, ALBERT plateaus the earliest, which may result from its parameter-sharing nature.

\section{World Knowledge Development}
\subsection{Dataset Statistics}
In our experiment of world knowledge, we only use 1-1 relations (P1376 and P36) and N-1 relations (the rest relations in Table~\ref{tab:rel_count}).
Among those relations, we only ask our model to predict object ([Y] in the template in Table~\ref{tab:rel_count}) that has only one token, following \citet{petroni2019language}.
From those relations, we report world knowledge that behaves differently during pretraining in Figure~\ref{fig:know}: we select the knowledge that can be learned during pretraining (e.g., P176), the knowledge that cannot be learned during the whole pretraining process (e.g., P140), the knowledge that was once learned and then forgotten after pretraining (e.g., P138), and knowledge that kept oscillating during pretraining (e.g., P407).
The statistics of all world knowledge evaluated are in listed in Table~\ref{tab:rel_count}.
\begin{table}[]
    \small
    \centering
    \begin{tabular}{ccc}
    \centering
        Type & Count & Template\\
        P140 & 471 & [X] is affiliated with the [Y] religion . \\
P103 & 975 & The native language of [X] is [Y] . \\
P276 & 954 & [X] is located in [Y] . \\
P176 & 946 & [X] is produced by [Y] . \\
P264 & 312 & [X] is represented by music label [Y] . \\
P30 & 975 & [X] is located in [Y] . \\
P138 & 621 & [X] is named after [Y] . \\
P279 & 958 & [X] is a subclass of [Y] . \\
P131 & 880 & [X] is located in [Y] . \\
P407 & 870 & [X] was written in [Y] . \\
P36 & 699 & The capital of [X] is [Y] . \\
P159 & 964 & The headquarter of [X] is in [Y] . \\
P17 & 930 & [X] is located in [Y] . \\
P495 & 909 & [X] was created in [Y] . \\
P20 & 952 & [X] died in [Y] . \\
P136 & 931 & [X] plays [Y] music . \\
P740 & 934 & [X] was founded in [Y] . \\
P1376 & 230 & [X] is the capital of [Y] . \\
P361 & 861 & [X] is part of [Y] . \\
P364 & 852 & The original language of [X] is [Y] . \\
P37 & 952 & The official language of [X] is [Y] . \\
P127 & 683 & [X] is owned by [Y] . \\
P19 & 942 & [X] was born in [Y] . \\
P413 & 952 & [X] plays in [Y] position . \\
P449 & 874 & [X] was originally aired on [Y] . 
    \end{tabular}
    \caption{Relations used.}
    \label{tab:rel_count}
\end{table}
\begin{figure*}[t]
    \centering
    \includegraphics[scale = 0.8]{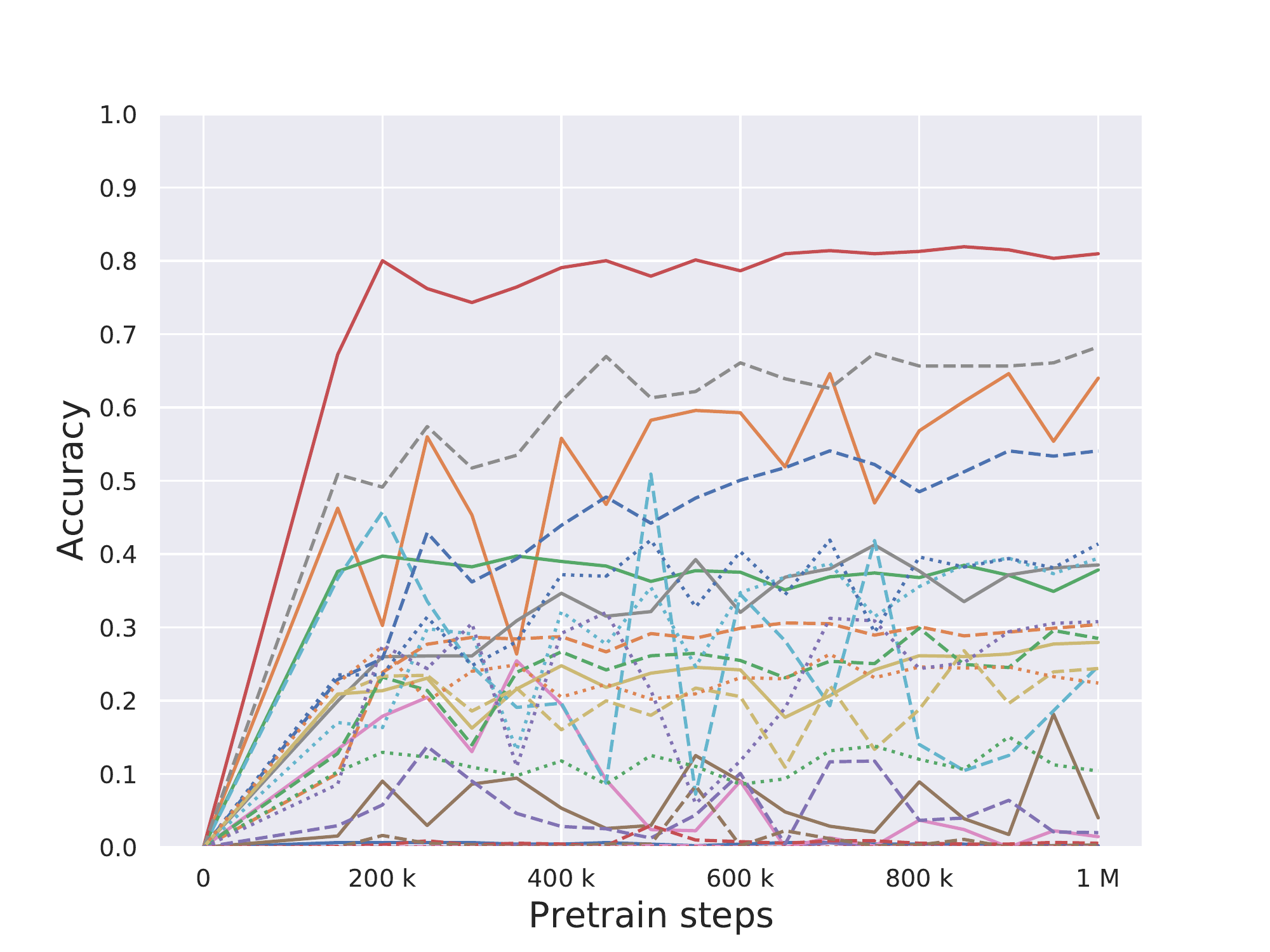}
    \includegraphics[scale = 0.9]{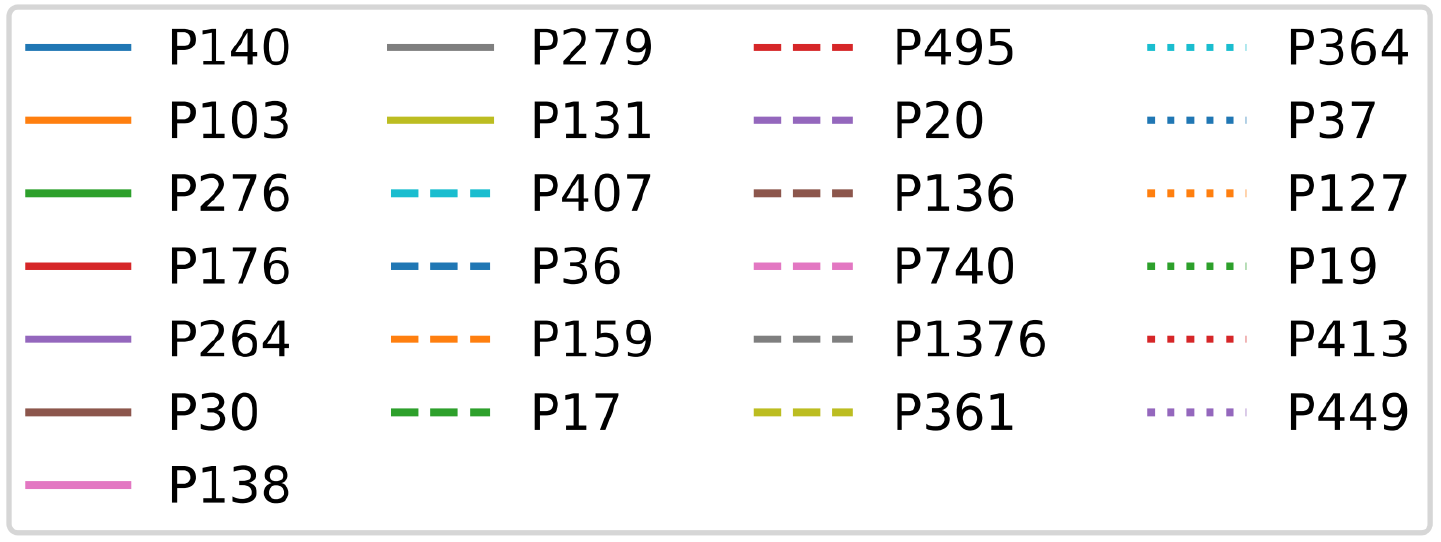}
    \caption{Prediction of all world knowledge during pretraining.}
    \label{fig:know_all}
\end{figure*}
\subsection{Qualitative Results and Complete World Knowledge Results}
We provide qualitative examples for Section~\ref{section: knowledge} in Table~\ref{tabel:text}.
We also provide the complete results of all world knowledge we use in Figure~\ref{fig:know_all}.
\label{app: qualitity}
\begin{table*}[t]
\centering
\begin{tabular}{ |p{1.2cm}|p{6cm}|p{6cm}|  }
 \hline
 \multicolumn{3}{|c|}{World Knowledge Prediction} \\
 \hline
 Relation & P38 & P176\\
 \hline
 Query & Nokia Lumia 800 was produced by [MASK].& Hamburg airport is named after [MASK].\\
 \hline
 Answer & Nokia & Hamburg \\
 \hline
 100K & the lumia 800 is produced by nokia.  & hamburg airport is named after it.\\
 200K &  nokia lu nokia 800 is produced by nokia.  & hamburg airport is named after hamburg.\\
 500K& nokia lumia 800 is produced by nokia. & hamburg airport is named after him.\\
 1M&nokia lumia 800 is produced by nokia. &hamburg airport is named after him.\\
 \hline
\end{tabular}
\caption{Example results of world knowledge evolution during pretraining. 
We can observe that model successfully predict the object in the Nokia example since 100K steps, and doesn't forget during the rest pretraining process. 
On the other hand, the model is only able to correctly predict Hamburg in the second example at 200K steps, and failed to predict at other pretrain steps.}
\label{tabel:text}
\end{table*}

\end{document}